\documentclass{article}
\usepackage{proceed2e}

\usepackage{mkolar_definitions}
\usepackage{color}
\let\chapter\section
\usepackage[ruled,linesnumbered]{algorithm2e}
\usepackage{algorithmic}
\usepackage[numbers]{natbib}
\usepackage{graphicx}

\usepackage[usenames,dvipsnames,svgnames,table]{xcolor}

\numberwithin{equation}{section}
\numberwithin{theorem}{section}

\newcommand{\removed}[1]{}

\newcommand{\estat}{\epsilon}
\newcommand{\eopt}{\epsilon_{\rm opt}}

\title{Distributed Multi-Task Learning with Shared Representation}

\author{} 

\author{ {\bf Jialei Wang} \\
Department of Computer Science \\
University of Chicago\\
Chicago, IL 60637 \\
\And
{\bf Mladen Kolar}  \\
Booth School of Business \\
University of Chicago\\
Chicago, IL 60637 \\
\And
{\bf Nathan Srebro}   \\
Toyota Technological Institute \\
at Chicago \\
Chicago, IL 60637 \\
}

\begin{document}

\maketitle

\begin{abstract}
  We study the problem of distributed multi-task learning with shared
  representation, where each machine aims to learn a separate, but
  related, task in an unknown shared low-dimensional subspaces,
  i.e.~when the predictor matrix has low rank.  We consider a setting
  where each task is handled by a different machine, with samples for
  the task available locally on the machine, and study
  communication-efficient methods for exploiting the shared structure.
\end{abstract}

\section{Introduction}

Multi-task learning is widely used learning framework in which similar
tasks are considered jointly for the purpose of improving performance
compared to learning the tasks separately \citep{caruana1997multitask}.
By transferring information between related tasks it is hoped that
samples will be better utilized, leading to improved generalization
performance.  Multi-task learning has been successfully applied, for
example, in natural language understanding \citep{collobert2011natural}, 
speech recognition \citep{seltzer2013multi}, 
remote sensing \citep{xue2007multi}, 
image classification \citep{lapin2014scalable},
spam filtering \citep{weinberger2009feature}, 
web search \citep{chapelle2010multi},
disease prediction \citep{zhou2013modeling},
and eQTL mapping \citep{kim2010tree}
among other applications.  

Here, we study multi-task learning in a distributed setting, where
each task is handled by a different machine and communication between
machines is expensive.  That is, each machine has access to data for a
different task and needs to learn a predictor for that task, where
machines communicate with each other in order to leverage the
relationship between the tasks.  This situation lies between a
homogeneous distributed learning setting \cite[e.g.][]{Shamir2014Distributed}, where all
machines have data from the same source distribution, and inhomogeneous consensus problems \cite[e.g.][]{Ram2010Distributed,Boyd2011Distributed,Balcan2012Distributed}
where the goal is to reach a single consensus predictor or iterate
which is the same on all machines.  The main argument for this setting
is that if each machine indeed has access to different data (e.g.~from
a different geographical region or different types of users), as in
the consensus problems studied by \citet{Balcan2012Distributed}, then we should allow
a different predictor for each distribution, instead of insisting on a
single consensus predictor, while still trying to leverage the
relationship and similarity between data distributions, as in
classical multi-task learning.  As was recently pointed out by
\citet{Wang2015Distributed}, allowing separate predictors for each task instead
of insisting on a consensus predictor changes the fundamental nature
of the distributed learning problem, allows for different
optimization methods, and necessitates a different analysis approach,
more similar to homogeneous distributed learning as studied by
\citet{Shamir2014Distributed}. 

The success of multi-task learning relies on the relatedness between
tasks.  While \citet{Wang2015Distributed} studied tasks related through shared
sparsity, here we turn to a more general, powerful and empirically
more successful model of relatedness, where the predictors for
different tasks lie in some (a-priori unknown) shared low-dimensional
subspace and so the matrix of predictors is of low rank
\citep{Ando2005framework,Amit2007Uncovering,yuan2007dimension,argyriou08convex}.
In a shared sparsity model, information from all tasks is used to
learn a subset of the input features which are then used by all tasks.
In contrast, in a shared subspace model, novel features, which are
linear functions of the input features, are learned.  The model can
thus be viewed as a two-layer neural network, with the bottom layer
learned jointly across tasks and the top layer task-specific.  Being
arguably the most complex multi-layer network that we can fully
analyze, studying such models can also serve as a gateway to using
deeper networks for learning shared representations.

Multi-task learning with a shared subspace is well-studied in a
centralized setting, where data for all tasks are on the same machine,
and some global centralized procedure is used to find a good predictor
for each task.  In such a situation, nuclear norm regularization is
often used to leverage the low rank structure
\citep[e.g.][]{argyriou08convex,Amit2007Uncovering} and learning
guarantees are known (\citep{Maurer2012Excess} and see also Section
\ref{sec:baseline}).  With the growth of modern massive data sets,
where tasks and data often too big to handle on a single machine, it
is important to develop methods also for the distributed setting.
Unfortunately, the distributed multi-task setting is largely
unexplored and we are not aware of any prior for on distributed
multi-task learning with shared subspaces.

In this paper we focus on methods with efficient communication
complexity (i.e.~with as small as possible communication between
machines), that can still leverage most of the statistical benefit of
shared-subspace multi-task learning.  Although all our methods are
also computationally tractable and can be implemented efficiently, we
are less concerned here with minimizing the runtime on each machine
separately, considering communication, instead, as the main bottleneck
and the main resource to be minimized \citep{Bekkerman2011Scaling}.
This is similar to the focus in distributed optimization approaches
such as ADMM \citep{Boyd2011Distributed} and DANE \citep{Shamir2013Communication} where optimization within each
machine is taken as an atomic step.

\begin{table*}[t]
\begin{small}
  \centering
  \begin{tabular}{|c||c|c|c|c|c|c|c|}
  \hline
  Approach  & Samples & Rounds & Communication & Worker Comp. & Master Comp. \\\hline\hline
  \texttt{Local} & $\frac{A^2}{\varepsilon^2}$ & $1$ & $0$  & ERM & $0$ \\ \hline
  \texttt{Centralize} & $\frac{A^2}{\estat^2} \left( \frac{r}{m} +
    \frac{r}{\tilde{p}} \right) $ & $1$ & $ \frac{A^2}{\estat^2} \left( \frac{r}{m} +
    \frac{r}{\tilde{p}} \right) $ & $0$ & Nuclear Norm Minimization \\
\hline \hline
  \texttt{ProxGD}  &  $\frac{A^2}{\estat^2} \left( \frac{r}{m} +
    \frac{r}{\tilde{p}} \right) $  &  $\frac{mH A^2}{\varepsilon}$  & $2 \cdot p$  & Gradient Comp. & SV Shrinkage  \\ \hline
  \texttt{AccProxGD}  &  $\frac{A^2}{\estat^2} \left( \frac{r}{m} +
    \frac{r}{\tilde{p}} \right) $  &  $\sqrt{\frac{mH A^2}{\varepsilon}}$  & $2 \cdot p$   & Gradient Comp. & SV Shrinkage   \\ \hline
  \texttt{ADMM}  &  $\frac{A^2}{\estat^2} \left( \frac{r}{m} +
    \frac{r}{\tilde{p}} \right) $  &  $\frac{m A^2}{\varepsilon}$  & $3 \cdot p$   &  ERM & SV Shrinkage \\ \hline
 \texttt{DFW}  &  $\frac{A^2}{\estat^2} \left( \frac{r}{m} +
    \frac{r}{\tilde{p}} \right) $  &  $\frac{mH A^2}{\varepsilon}$  & $2 \cdot p$  & Gradient Comp. & Leading SV Comp.   \\ 
 \hline \hline
   \texttt{DGSP}  &  $-$  &  $\frac{mH A^2}{\varepsilon}$   & $2 \cdot p$   & ERM & Leading SV Comp. \\ \hline
   \texttt{DNSP}     &  $-$  &  $-$  & $2 \cdot p$   & ERM & Leading SV Comp.  \\
   \hline \hline
  \end{tabular}
  \caption{\small Summary of resources required by different approaches to
    distributed multi-task learning with shared representations (for squared loss), in units of vector operations/communications, ignoring log-factors.}
  \label{table:summary}
\end{small}
\end{table*}

\paragraph{Contribution} The main contributions of this article are:
\begin{itemize}
\itemsep0em 
\item Present and formalize the shared-subspace multi-task learning
  \citep{argyriou08convex} in the novel distributed multitask setting,
  identifying the relevant problems and possible approaches.  We
  analyze two baselines, several representative first-order
  distributed optimization methods, with careful sample and
  communication complexity analysis.

\item We proposed and analyzed two subspace pursuit approaches which
  learns the shared representation in a greedy fashion, which leverage
  the low-dimensional predictive structure in a communication
  efficient way.

\item We conducted comprehensive experimental comparisons of the
  discussed approaches on both simulated and real datasets, where we
  demonstrated that the proposed approaches are more communication
  efficient than first-order convex optimization methods.
\end{itemize}

Table \ref{table:summary} summarized the approaches studied in this
paper, which will be discussed in detail in the following sections.

\paragraph{Homogeneous, Inhomogeneous and Multi-Task Distributed Learning.}
We briefly review the relationship between homogeneous, inhomogeneous
and multi-task learning, as recently presented by \citet{Wang2015Distributed}.

A typical situation considered in the literature is one in which
data on different machines are all drawn i.i.d from the same source
distribution. In this setting, tasks on different machines are all the
same, which should be taken advantage of in optimization
\cite{Shamir2013Communication}. Furthermore, as each machine has
access to samples from the source distribution it can perform
computations locally, without ever communicating with other machines.
While having zero communication cost, this approach does not compare
favorably with the centralized approach, in which all data are
communicated to the central machine and used to obtain one predictor,
when measured in terms of statistical efficiency.  The goal in this
setting is to obtain performance close to that of the centralized
approach, using the same number of samples, but with low communication
and computation costs \cite{Shamir2014Distributed,
  Jaggi2014Communication, Zhang2013Information,
  Zhang2012Communication, lee2015communication}.  Another setting
considered in the distributed optimization literature is that of
consensus optimization. Here each machine has data from a different
distribution and the goal is to find one vector of coefficients that
is good for all the separate learning or optimization problems
\cite{Boyd2011Distributed, Ram2010Distributed, Balcan2012Distributed}.
The difficulty of consensus problems is that the local objectives
might be rather different, and, as a result, one can obtain lower
bounds on the amount of communication that must be exchanged in order
to reach a joint optimum.
  
In this paper we suggest a novel setting that combines aspects of the
above two settings.  On one hand, we assume that each machine has a
different source distributions $\Dcal_j$, corresponding to a different
task, as in consensus problems.  For example, each machine serves a
different geographical location, or each is at a different hospital or
school with different characteristics.  But if indeed there are
differences between the source distributions, it is natural to learn
different predictors $\wb_j$ for each machine, so that $\wb_j$ is good
for the distribution typical to that machine.  In this regard, our
distributed multi-task learning problem is more similar to
single-source problems, in that machines could potentially learn on
their own given enough samples and enough time. Furthermore,
availability of other machines just makes the problem easier by
allowing transfer between the machine, thus reducing the sample
complexity and potentially runtime.  The goal, then, is to leverage as
much transfer as possible, while limiting communication and runtime.
As with single-source problems, we compare our method to the two
baselines, where we would like to be much better than the local
approach, achieving performance nearly as good as the centralized
approach, but with minimal communication and efficient runtime.

\section{Setting, Formulation and Baselines}
\label{sec:baseline}

We consider a setting with $m$ tasks, each characterized by a source
distribution $\Dcal_j(\Xb,Y)$ over feature vectors $\Xb\in\RR^p$ and
associated labels $Y$, and out goal is to find linear predictors
$\wb_1,\ldots,\wb_m \in \RR^p$ minimizing the overall expected loss (risk)
across tasks:
\begin{equation}
  \label{eq:avg_risk}
  \Lcal(W) = 
  \frac{1}{m} \sum_{j=1}^m \EE_{
    (\Xb_j,Y_j) \sim \Dcal_j} \sbr{ \ell(\wb_j^T\Xb_j,Y_j)}  ,
\end{equation}
where for convenience we denote $W\in\RR^{p \times m}$ for the matrix
with columns $\wb_i$, and $\ell(\cdot,\cdot)$ is some specified
instantaneous loss function.

In the learning setting, we cannot observe $\Lcal(W)$ directly and
only have access to i.i.d.~sample $\{\xb_{ji}, y_{ji}\}_{i=1}^{n_j}$
from each distribution $\Dcal_j$, $j=1,\ldots, m$.  For simplicity of
presentation, we will assume that $n_j = n$, $j=1,\ldots, m$,
throughout the paper.  We will denote the empirical loss $\Lcal_n(W) =
\frac{1}{m}\sum_{j=1}^m \Lcal_{nj}(\wb_j)$ where
\[
\Lcal_{nj}(\wb_j) = \frac{1}{n} \sum_{i=1}^n \ell(\wb_j^T\xb_{ji},y_{ji})
\]
is the local (per-task) empirical loss.

We consider a distributed setting, where each task is handled on one
of $m$ separate machines, and each machine $j$ has access only to the
samples drawn from $\Dcal_j$.  Communication between the machines is
by sending real-valued vectors.  Our methods work either in a
broadcast communication setting, where at each iteration each machine
sends a vector which is received by all other machines, or in a
master-at-the-center topology where each machine sends a vector to the
master node, whom in turn performs some computation and broadcasts
some other vectors to all machines.  Either way, we count to total
number of vectors communicated.

As in standard agnostic-PAC type analysis, our
goal will be to obtain expected loss $\Lcal(W)$ which is not much
larger then the expected loss of some (unknown) reference
predictor\footnote{Despite the notation, $W^*$ need {\em not} be the
  minimizer of the expected loss.  We can think of it as the minimizer
  inside some restricted hypothesis class, though all analysis and
  statements hold for any chosen reference predictor $W^*$.} $W^*$,
and we will measure the {\em excess error} over this goal.  To allow
obtaining such guarantees we will assume:
\begin{assumption}
\label{assum:lipschitz}
The loss function $\ell(\cdot)$ is $1$-Lipschitz and
bounded\footnote{This is only required for the high probability bounds.} by $1$, be twice differentiable and $H$-smooth, that is 
\[
| \ell'(a,c) - \ell'(b,c)| \leq H |a - b|, \qquad \forall a,b,c \in \RR.
\]
All the data points are bounded by unit length,
i.e.
\[
\norm{\xb_{ji}}_2 \leq 1, \forall i,j,
\]
and the reference predictors have bounded norm:
\[
\max_{j \in [m]} \norm{\wb_j^*}_2^2 \leq A^2
\]
for some $A < \infty$.
\end{assumption}

The simplest approach, which we refer to as \texttt{Local}, is to
learn a linear predictor on each machine independently of other
machines.  This single task learning approach ignores the fact that
the tasks are related and that sharing information between them could
improve statistical performance.  However, the communication cost for
this procedure is zero, and with enough samples it can still drive the
excess error to zero. However, compared to procedures discussed later,
sample complexity (number of samples $n$ required to achieve small
excess error) is larger.  A standard Rademacher complexity argument
\citep{bartlett2003rademacher} gives the following generalization
guarantee, which is an extension of Theorem 26.12 in
\citet{Shalev-Shwartz2014Understanding}.
\begin{proposition}
  Suppose Assumption \ref{assum:lipschitz} holds. Then with probability at
  least $1 - \delta$, 
 \begin{align*}
 \Lcal(\hat W_{\rm local}) - \Lcal(W^*) \leq \frac{2A}{\sqrt{n}} + \sqrt{\frac{2 \ln (2m/\delta)}{n}},
 \end{align*}
 where $\hat W_{\rm local} = [\hat\wb_1,\ldots,\hat\wb_m]$ with
 $\hat \wb_j = \arg \min_{\norm{\wb}\leq A} \Lcal_{nj}(\wb)$.
\end{proposition}
That is, in order to ensure $\estat$ excess error, we need
\begin{align*}
n = \Ocal \rbr{ \frac{A^2}{\estat^2} }
\end{align*}
samples from each task.

At the other extreme, if we ignore all communication costs, and,
e.g.~communicate all data to a single machine, we can significantly
leverage the shared subspace.  To understand this, we will first need
to introduce two assumptions: one about the existence of a shared
subspace (i.e.~that the reference predictor is indeed low-rank), and
the other about the spread of the data:

\begin{assumption}
\label{assum:rank} ${\rm rank}(W^*) \leq r$
\end{assumption}

\begin{assumption}
\label{assum:spectral}
There is a constant $\tilde{p}$, such that
\begin{align*}
\bignorm{ \frac{1}{m} \sum_{j=1}^m \EE_{(\Xb_j,Y_j)\sim \Dcal_j} \sbr{\Xb_{j} \Xb_{j}^T} }_2 \leq \frac{1}{\tilde{p}}.
\end{align*}
\end{assumption}
Since the data is bounded, we always have $1 \leq \tilde{p} \leq p$, with $\tilde{p}$
being a measure of how spread out the data is in different direction.
A value of $1=\tilde{p}$ indicates the data is entirely contained in a
one-dimensional line.  In this case, the predictor matrix will also
always be rank-one, imposing a low-rank structure is meaningless and
we can't expect to gain from it. However, when $\tilde{p}$ is close to
$p$, or at least high, the data is spread in many directions and the
low-rank assumption is meaningful.  We can think of $\tilde{p}$ as the
``effective dimensionality'' of the data, and hope to gain when
$r \ll \tilde{p}$.

With these two assumptions in hand, we can think of minimizing the
empirical error subject to a rank constraint on $W$.  This is a hard
and non-convex optimization task, but we can instead use the nuclear
norm (aka trace-norm) $\norm{W}_*$ as a convex surrogate for the rank.
This is because if Assumptions \ref{assum:lipschitz} and
\ref{assum:rank} hold, then we also have:
\begin{align}
\norm{W^*}_* \leq \sqrt{rm} A.
\end{align}
With this in mind, we can define the following {\em centralized}
predictor:
\begin{align}
\hat W_{\rm centralize} = \arg\min_{\norm{W}_*\leq \sqrt{rm}A}
\Lcal_n(W) \label{eq:nnmtl}
\end{align}
which achieves the improved
excess error guarantee:
\begin{proposition}(Theorem 1 in \citet{Maurer2012Excess}) 
  Suppose Assumptions \ref{assum:lipschitz}, \ref{assum:rank} and \ref{assum:spectral} hold.
  Then  with probability at least $1 - \delta$,
\begin{align*}
\Lcal(\hat W_{\rm centralize})  \leq& \Lcal(W^*) + \sqrt{\frac{2 \ln(2/\delta)}{nm}} \\
 &+2 \sqrt{r} A \rbr{ \sqrt{\frac{1}{\tilde{p} n}} + 5 \sqrt{\frac{\ln(mn) + 1}{mn}} }  
\end{align*}
\end{proposition}
The sample complexity per task, up to logarithmic factors, is thus only:
\begin{align*}
n = \tilde{\Ocal}\left(\frac{A^2}{\estat^2} \left( \frac{r}{m} +
    \frac{r}{\tilde{p}} \right) \right)
\end{align*}
When $\tilde{p} \gg m$, this is a reduction by a factor of $r/m$.
That is, it is as if we needed to only learn $r$ linear predictors
instead of $m$.

The problem is that a naive computation of $\hat W_{\rm centralize}$
requires collecting all data on a single machine, i.e.~communicating
$O(n)=\tilde{\Ocal}\left(\frac{A^2}{\estat^2} \left( \frac{r}{m} +
    \frac{r}{\tilde{p}} \right) \right)$ samples per machine.  In the
next Sections, we aim at developing methods of approximating $\hat
W_{\rm centralized}$ using communication efficient methods, or
computing an alternate predictor with similar statistical properties
but using much less communication.

\section{Distributed Convex Optimization}
\label{sec:first-order}

In this section, we study how to obtain the sharing benefit of the
centralized approach using distributed convex optimization techniques,
while keeping the communication requirements at low.

To enjoy the benefit of nuclear-norm regularization while avoid heavy communication cost of \texttt{Centralize}, a flexible strategy is to solve the convex objective \eqref{eq:nnmtl} via distributed optimization techniques.  Let $W^{(t)}$ be the solution at $t$-iteration for some iterative distributed optimization algorithm for the following constrained objective:
\begin{align}
\min_{\norm{W}_* \leq \sqrt{rm} A} \Lcal_n(W).
\label{eqn:constrained_objective}
\end{align}
By the generalization error decompsition \citep{bousquet2008tradeoffs}, 
\begin{align*}
\Lcal(W^{(t)}) - \Lcal(W^*) \leq & 2 \estat + \eopt,
\end{align*}

\removed{
\begin{align*}
\Lcal(W^{(t)}) - \Lcal(W^*) =& \underbrace{\Lcal(W^{(t)}) -  \Lcal_n(W^{(t)})}_{\leq \estat} \\
&+ \underbrace{\Lcal_n(W^{(t)}) - \Lcal_n(\hat W)}_{\leq \eopt} \\
&+ \underbrace{\Lcal_n(\hat W) - \Lcal_n(W^*)}_{\leq 0} \\
&+ \underbrace{\Lcal_n(W^*) - \Lcal(W^*)}_{\leq \estat}  \\
\leq & 2 \estat + \eopt,
\end{align*}}

\removed{
{\bf TODO:} $\estat$ as introduced on page 1 is somewhat different. If I understand correctly
 the meaning here is 
\[
\sup_W \abr{\Lcal_n(W) - \Lcal(W)} \leq \estat.
\]
If that is the case, then you should modify the definition in the intro which states
\[
\Lcal(\hat W) \leq \Lcal(W^*) + \estat
\]

{\bf end todo.}
}

Suppose $W^{(t)}$ satisfying
$\Lcal_n(W^{(t)}) \leq \Lcal_n(\hat W) + \Ocal(\eopt)$ with
$\eopt = \Ocal(\estat)$. Then $W^{(t)}$ will have the generalization
error of order $\Ocal(\estat)$.  Therefore in order to study the
generalization performance, we will study how the optimization error
decreases as the function of the number of iterations $t$.

\paragraph{Constrained vs Regularized Objective} Note that the constrained objective \eqref{eqn:constrained_objective} is equivalent to the following regularized objective with a proper choice of $\lambda$:
\begin{align}
\min_{W} \Lcal_n(W) + \lambda \norm{W}_*.
\label{eqn:regularized_objective}
\end{align}
Though they are equivalent, specific optimization algorithms might sometimes be more suitable for one particular type of objectives \footnote{e.g. ADMM for regularized objective and Frank-Wolfe for constrained objective. Gradient descent methods can be adopted for both, leads to proximal and projected methods, respectively.}. For convenience in the following discussion we didn't distinguish between these two formulations.

\subsection{Distributed Proximal Gradient}

Maybe the simplest distributed optimization algorithm for
\eqref{eqn:regularized_objective} is the proximal gradient descent. It is not hard to
see that computation of the gradient $\nabla \Lcal_n(W)$ can be
easily done in a distributed way as the losses are decomposable
across machines:
\begin{align*}
\nabla \Lcal_n(W) = 
\begin{bmatrix}
\nabla \Lcal_{n1}(\wb_1), \ldots, \nabla \Lcal_{nm}(\wb_m)
\end{bmatrix}
\end{align*}
where
\[
\nabla \Lcal_{nj}(\wb_j) = \frac{1}{nm} \sum_{i=1}^n \ell'(\dotp{\wb_j}{\xb_{ji}}, y_{ji}) \xb_{ji}.
\]
Thus each machine $j$  needs to compute the gradient $\nabla \Lcal_{nj}(\wb_j)$ 
on the local dataset and send it to the master. The master concatenates
the gradient vectors to form the gradient matrix $\nabla \Lcal_n(W)$.
Finally, the master computes the proximal step
\begin{align}
W^{(t+1)} = \arg\min_{W} & \norm{W - (W^{(t)} - \eta \nabla \Lcal_n(W^{(t)}))}_F^2 \nonumber \\
&+ \lambda \norm{W}_*,
\label{eq:prox_op}
\end{align}
which has the following closed form solution \citep{cai2010singular}: let
$W^{(t)} - \eta \nabla \Lcal_n(W^{(t)}) = U\Sigma V^T$ be the SVD of
$W^{(t)} - \eta \nabla \Lcal_n(W^{(t)})$, then $W^{(t+1)} = U \rbr{ \Sigma - 0.5 \lambda I }_+ V^T$
with $(x)_+ = \max\{0, x\}$ applied element-wise.  

The algorithm is summarized in Algorithm \ref{alg:prox_gd} (in Appendix), which has
well established convergence rates \citep{Bach2011Optimization}:
\begin{align*}
\Lcal_n(W^{(t)}) - \Lcal_n(\hat W) \leq \frac{ m H A^2}{2 t}.
\end{align*}
To obtain $\varepsilon$-generalization error, the distributed proximal gradient descent
requires $\Ocal \rbr{\frac{m H A^2}{\varepsilon}}$ rounds of communication, with a total
$\Ocal \rbr{\frac{m H A^2 p}{\varepsilon}}$ bits communications per machine.\\

\subsection{Distributed Accelerated Gradient}

It is also possible to use Nesterov's acceleration idea
\citep{nesterov1983method} to improve the convergence of the proximal
gradient algorithm from $\Ocal \rbr{\frac{1}{t}}$ to
$\Ocal \rbr{\frac{1}{t^2}}$ \citep{Ji2009accelerated}.  Using the
distributed accelerated proximal gradient descent, one needs
$\Ocal \rbr{ \sqrt{\frac{m H A^2}{\varepsilon} }}$ rounds of
communication with a total
$\Ocal \rbr{ \sqrt{\frac{m H A^2 }{\varepsilon}} \cdot p}$ bits
communicated per machine to achieve $\varepsilon$-generalization
error.  The algorithm is summarized in Algorithm
\ref{alg:acc_prox_gd} (in Appendix), where the master maintains two sequences: $W$ and
$Z$. First, a proximal gradient update of $W$ is done based on $Z$
\begin{align}
W^{(t+1)} = \arg\min_{Z} & \norm{Z - (Z^{(t)} - \eta \nabla \Lcal_n(Z^{(t)}))}_F^2 \nonumber \\
&+ \lambda \norm{Z}_*
\label{eq:prox_op_z}
\end{align}
and then $Z$ is updated based on a combination of the current $W$ and
the difference with previous $W$
\begin{align}
Z^{(t+1)} = W^{(t+1)} + \gamma_t (W^{(t+1)} - W^{(t)}).
\label{eq:acc_prox} 
\end{align}

\paragraph{ADMM and DFW} We also discuss the implementation and guarantees for two other popular optimization methods: ADMM and Frank-Wolfe, which are presented in the Appendix \ref{sec:admm} and \ref{sec:dfw}.

\section{Greedy Representation Learning}
\label{sec:greedy}

\begin{algorithm}[t]
\SetAlgoLined
\For{$t=1, 2, \ldots$}{
\underline{\textbf{Workers:}}\\
\For{$j=1, 2, \ldots, m$}{
Each worker compute the its gradient direction $ \nabla \Lcal_{nj}(\wb_j^{(t)}) $, and send it to the master
}
\If{Receive $\ub$ from the master}{
Update the projection matrix $U = [U ~ \ub]$;\\
Solve the projected ERM problem: $\vb_j = \arg\min_{\vb_j} \Lcal_{nj}(U\vb_j)$;\\
Update $\wb^{(t+1)}_j = U \vb_j$.
}
\underline{\textbf{Master:}}\\
\If{Receive $\nabla \Lcal_{nj}(\wb_j^{(t)}) $ from all workers}{
Concatenate the gradient vectors, and compute the largest singular vectors: $(\ub,\vb) = \textsf{SV}(\nabla \Lcal_{n}(W^{(t)}))$;\\
Send $\ub$ to all workers.
}
}
\caption{\texttt{DGSP}: Distributed Gradient Subspace Pursuit.}
\label{alg:dgsp}
\end{algorithm}

In this section we propose two distributed algorithms which select the
subspaces in a greedy fashion, instead of solving the nuclear norm regularized convex program.

\subsection{Distributed Greedy Subspace Pursuit}
\vspace{-0.2 cm}

Our greedy approach is inspired by the methods used for sparse signal
reconstruction \citep{tropp2004greed,Shalev-Shwartz2010Trading}.
Under the assumption that the optimal model $W^*$ is low-rank, say
rank $r$, we can write $W^*$ as a sum of $r$ rank-$1$ matrices:
\begin{align*}
W^* = \sum_{i=1}^r a_i \ub_i \vb_i^T = UV^T,
\end{align*}
where $a_i \in \RR, \ub_i \in \RR^{p}, \vb_i \in \RR^{m}$, and
$\norm{\ub_i}_2 = \norm{\vb_i}_2 = 1$. In the proposed approach, the projection matrix $U$ is
learned in a greedy fashion. At every iteration, a new one-dimensional
subspace is identified that leads to an improvement in the
objective. This subspace is then included into the existing projection
matrix. Using the new expanded projection matrix as the current
feature representation, we refit the model to obtain the coefficient
vectors $V$. In the distributed setting, there is a master that
gathers local gradient information from each task. Based on this
information, it then computes the subspace to be added to the
projection matrix and sends it to each machine.  The key step in the
distributed greedy subspace pursuit algorithm is the addition of the
subspace.  One possible choice is the principle component of the
gradient direction; after the master collected the gradient matrix
$\nabla \Lcal_{n}(W^{(t)})$, it computes the top left and right
singular vectors of $\nabla \Lcal_{n}(W^{(t)})$. Let
$(\ub,\vb) = \textsf{SV}(\nabla \Lcal_n(W^{(t)}))$ be the largest
singular vectors of $\nabla \Lcal_n(W^{(t)})$. The left singular
vector $\ub$ is used as a new subspace to be added to the projection
matrix $U$.  This vector is sent to each machine, which then
concatenate it to the projection matrix and refit the model with the
new representation. Algorithm~\ref{alg:dgsp} details the steps. 

\removed{
\mcomment{We have to many acronyms in the paper. It would be easier for readers if we used full name instead.}
}

Distributed gradient subspace pursuit (\texttt{DGSP}), detailed in
Algorithm~\ref{alg:dgsp}, creates subspaces that are orthogonal to
each other, as shown in the following proposition which is proved in Appendix \ref{sec:orthogonal}:
\begin{proposition}
At every iteration of Algorithm \ref{alg:dgsp}, the columns of $U$ are orthonormal. 
\label{prop:orthogonal}
\end{proposition}

Both the distributed gradient subspace pursuit and the distributed
Frank-Wolfe use the leading singular vector of the gradient matrix
iteratively. Moreover, leading
singular vectors of the gradient matrix have been used in greedy
selection procedures for solving low-rank matrix learning problems
\citep{Shalev-Shwartz2011Large, wang2015orthogonal}. However, \texttt{DGSP} utilize the learned subspace in a very different way: \texttt{GECO} \citep{Shalev-Shwartz2011Large} re-fit the
low-rank matrix under a larger subspace which is spanned by all left and right singular vectors; while \texttt{OR1MP}
\citep{wang2015orthogonal} only adjust the linear combination
parameters $\{a_i\}_{i=1}^r$ of the rank-1 matrices. The \texttt{DGSP} algorithm do not restrict on the joint subspaces $\{\ub_i \vb_i^T\}$, but focused on the low-dimensional subspace induced the projection matrix $U$, and estimate the task specific predictors $V$ based on the learned representation.

Next, we present convergence guarantees for the distributed gradient
subspace pursuit.  First, note that the smoothness of $\ell(\cdot)$
implies the smoothness property for any rank-1 update.
\begin{proposition}
  Suppose Assumption~\ref{assum:lipschitz} holds. Then for any $W$ and unit length vectors 
  $\ub \in \RR^p$ and $\vb \in \RR^m$, we have
\begin{align*}
\Lcal_n(W + \eta \ub \vb^T) \leq  \Lcal_n(W) + \ub^T \nabla \Lcal_n(W) \vb + \frac{H \eta^2}{2}.
\end{align*} 
\label{prop:smoothness}
\end{proposition}
We defer the proof in Appendix \ref{sec:proofprop}. The following theorem states the number of iterations needed for the
distributed gradient subspace pursuit to find an
$\varepsilon$-suboptimal solution.
\begin{theorem}
\label{thm:dgsp}
Suppose Assumption~\ref{assum:lipschitz} holds. Then the distributed
gradient subspace pursuit finds $W^{(t)}$ such that
$ \Lcal_n(W^{(t)}) \leq \Lcal_n(W^*) + \varepsilon$ 
\removed{
\mcomment{Should
  this be $\hat W$ instead of $W^*$?} 
  }
   when
\[
t \geq \left\lceil \frac{4 H m A^2}{\varepsilon} \right\rceil.
\]
\end{theorem}

We defer the proof in Appendix \ref{sec:proofthm}. Theorem \ref{thm:dgsp} tells us that for the distributed gradient
subspace pursuit requires $\Ocal \rbr{\frac{mHA^2}{\varepsilon}}$
iterations to reach $\epsilon$ accuracy. Since each iteration requires
communicating $p$ number, the communication cost per machine is
$\Ocal \rbr{\frac{mHA^2}{\varepsilon} \cdot p}$.  In some applications
this communication cost might be still too high and in order to
improve it we will try to reduce the number of rounds of
communication.  To that end, we develop a procedure that utilizes the
second-order information to improve the convergence. Algorithm
\ref{alg:dnsp} describes the Distributed Newton Subspace Pursuit
algorithm (\texttt{DNSP}). Note that distributed optimization with second-order information have
been studied recently to achieve communication efficiency
\citep{Shamir2013Communication,zhang2015communication}.

Compared to the gradient based methods,
the \texttt{DNSP} 
algorithm uses second-order information to find subspaces to work with.
At each iteration, each machine computes the Newton direction 
\begin{small}
\begin{align*}
\Delta \Lcal_{nj}(\wb_j) =& [\nabla^2 \Lcal_{nj}(\wb_j)]^{-1} \nabla
\Lcal_{nj}(\wb_j) \\
=& \sbr{\frac{1}{mn} \sum_{i=1}^{n} \ell''(\wb_j^T \xb_{ji},y_{ji}) \xb_{ji} \xb_{ji}^T }^{-1}  \nabla \Lcal_{nj}(\wb_j),
\end{align*}
\end{small}based on the current solution and  sends it to the master. The master computes
the overall Newton direction by concatenating the
Newton direction for each task
\begin{align*}
\Delta \Lcal_n(W) = 
[\Delta \Lcal_{n1}(\wb_1),\Delta \Lcal_{n2}(\wb_2),\ldots,\Delta \Lcal_{nm}(\wb_m)]
\end{align*}
and computes the top singular vectors of $\Delta \Lcal_n(W)$. The top
left singular vector $\ub$ is is sent back to every machine, which
is then concatenated to the current projection matrix. Each machine re-fits the
predictors using the new representation. Note that at every iteration
a Gram-Schmidt step is performed to ensure that the learned basis are
orthonormal. 

\texttt{DNSP} is a Newton-like method which uses second-order information, thus its generic analysis is not immediately apparent. However empirical results in the next section illustrate good performance of the proposed \texttt{DNSP}.

\removed{
{\bf TODO:} Add some discussion about this procedure. Mention that it is hard to study theoretically, but 
results in the next section illustrate good performance. }

\begin{figure*}[t]
\begin{center}
\includegraphics[width=0.33 \textwidth]{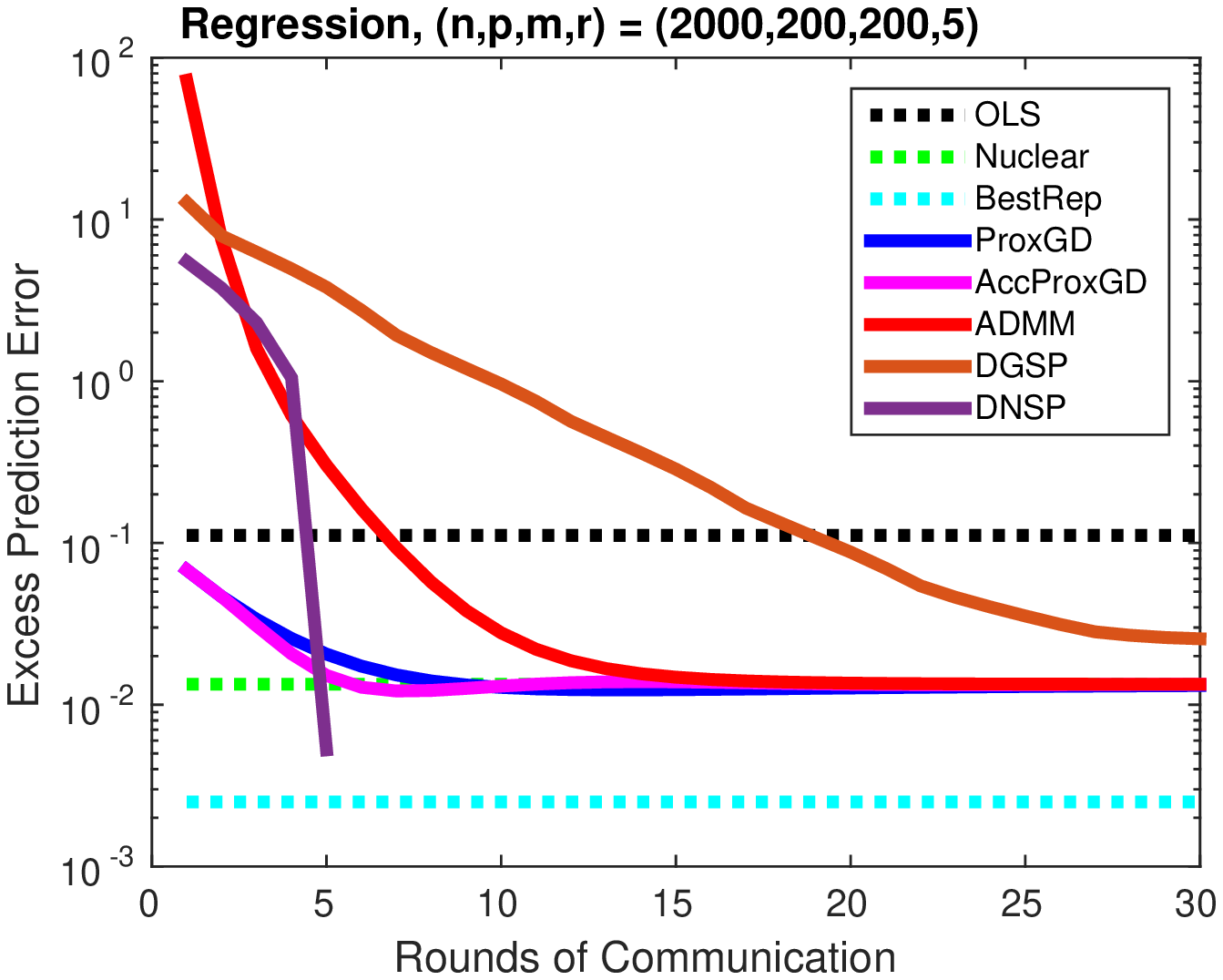}%
\includegraphics[width=0.33 \textwidth]{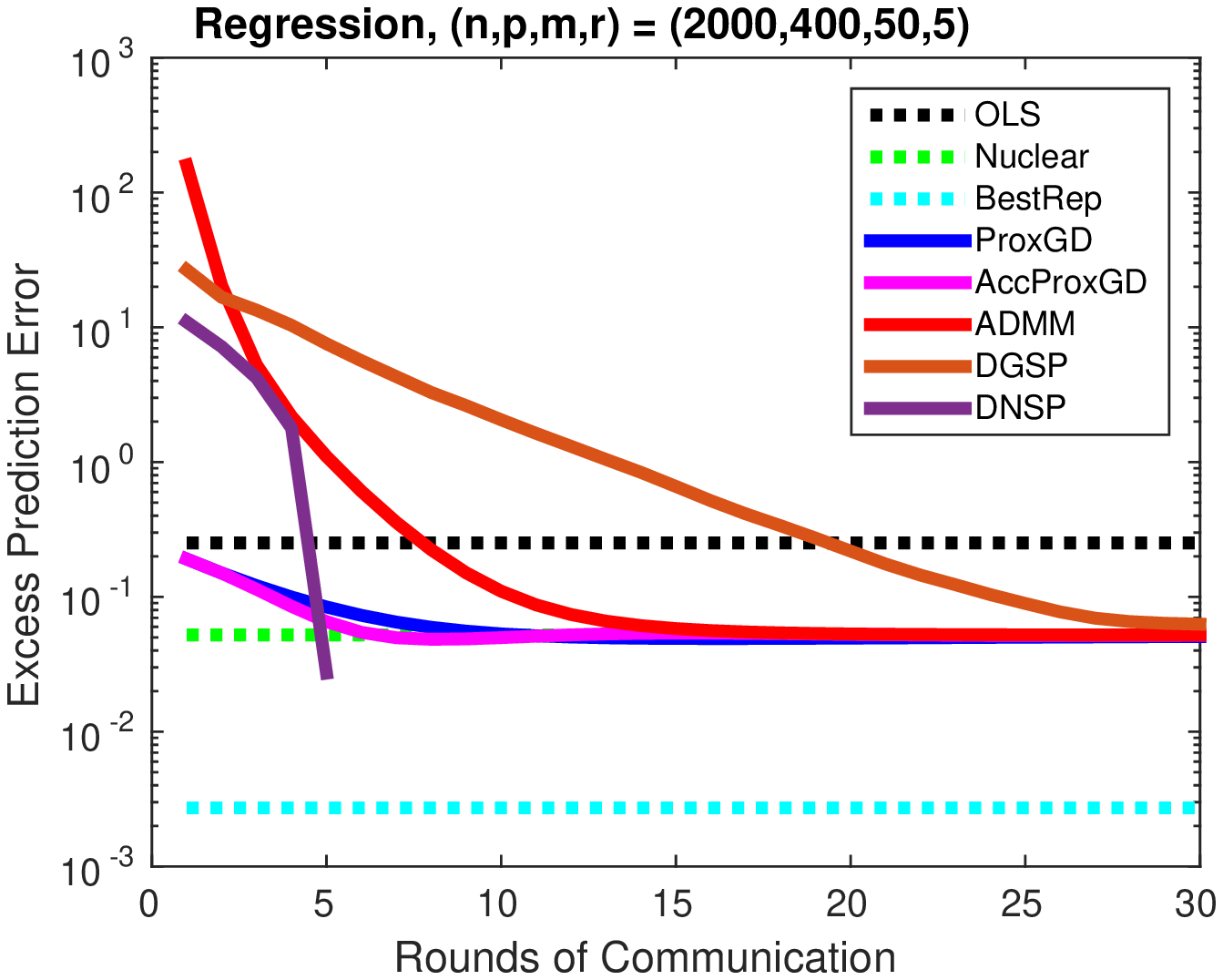}%
\includegraphics[width=0.33 \textwidth]{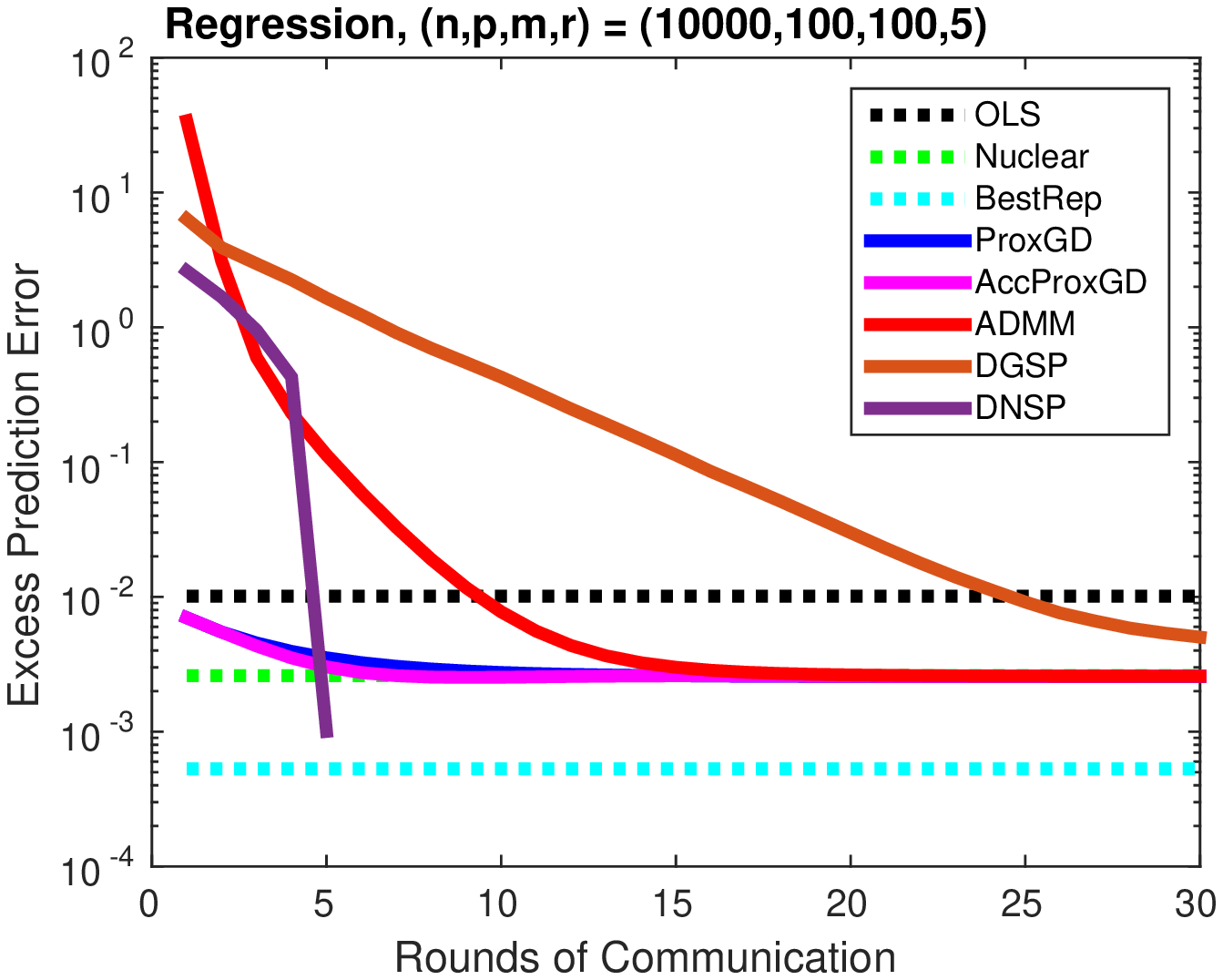}%
\end{center}
\caption{Excess prediction error for
  multi-task regression.}
\label{fig:simulation_regression}
\end{figure*}

\begin{figure*}[t]
\begin{center}
\includegraphics[width=0.33 \textwidth]{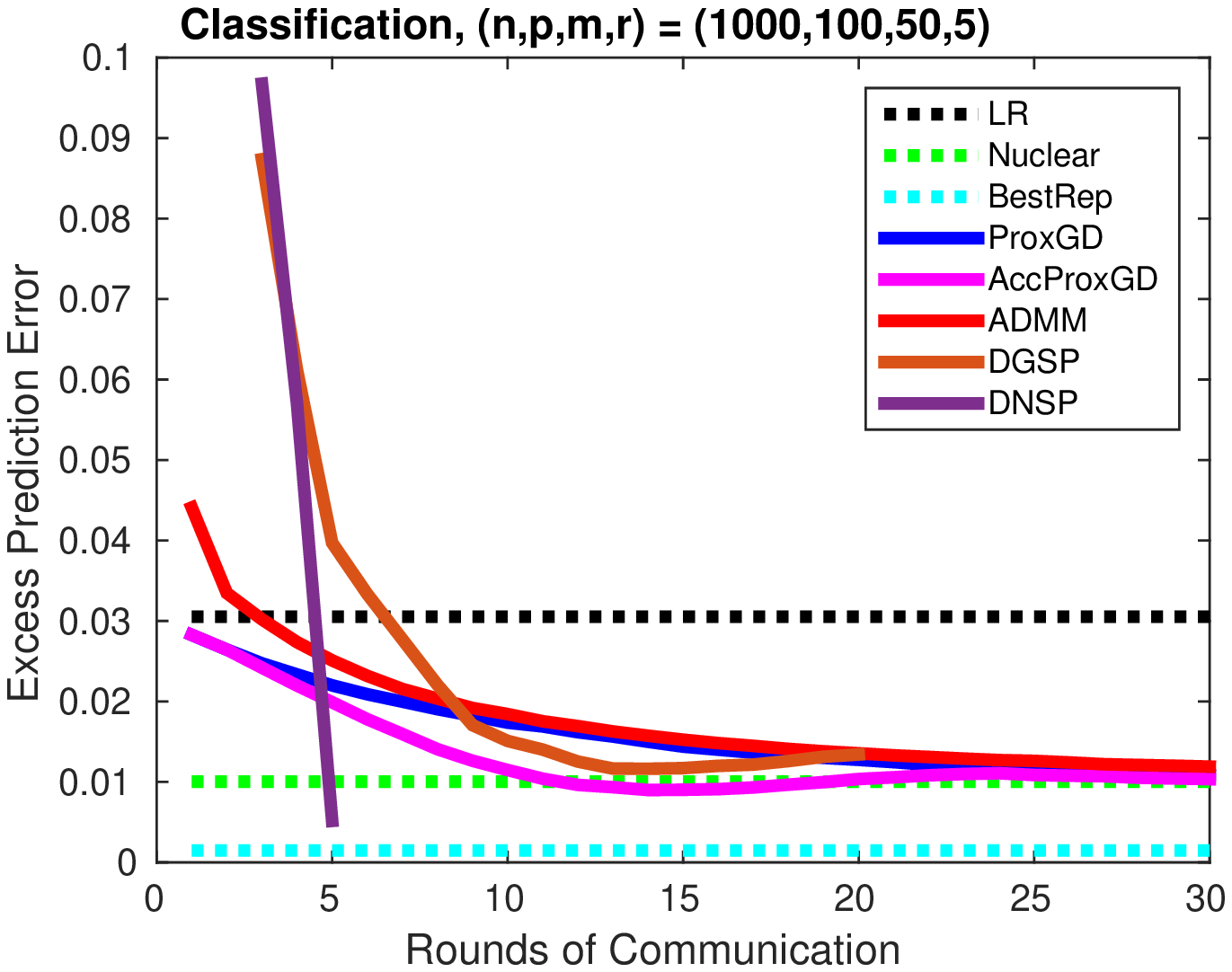}%
\includegraphics[width=0.33 \textwidth]{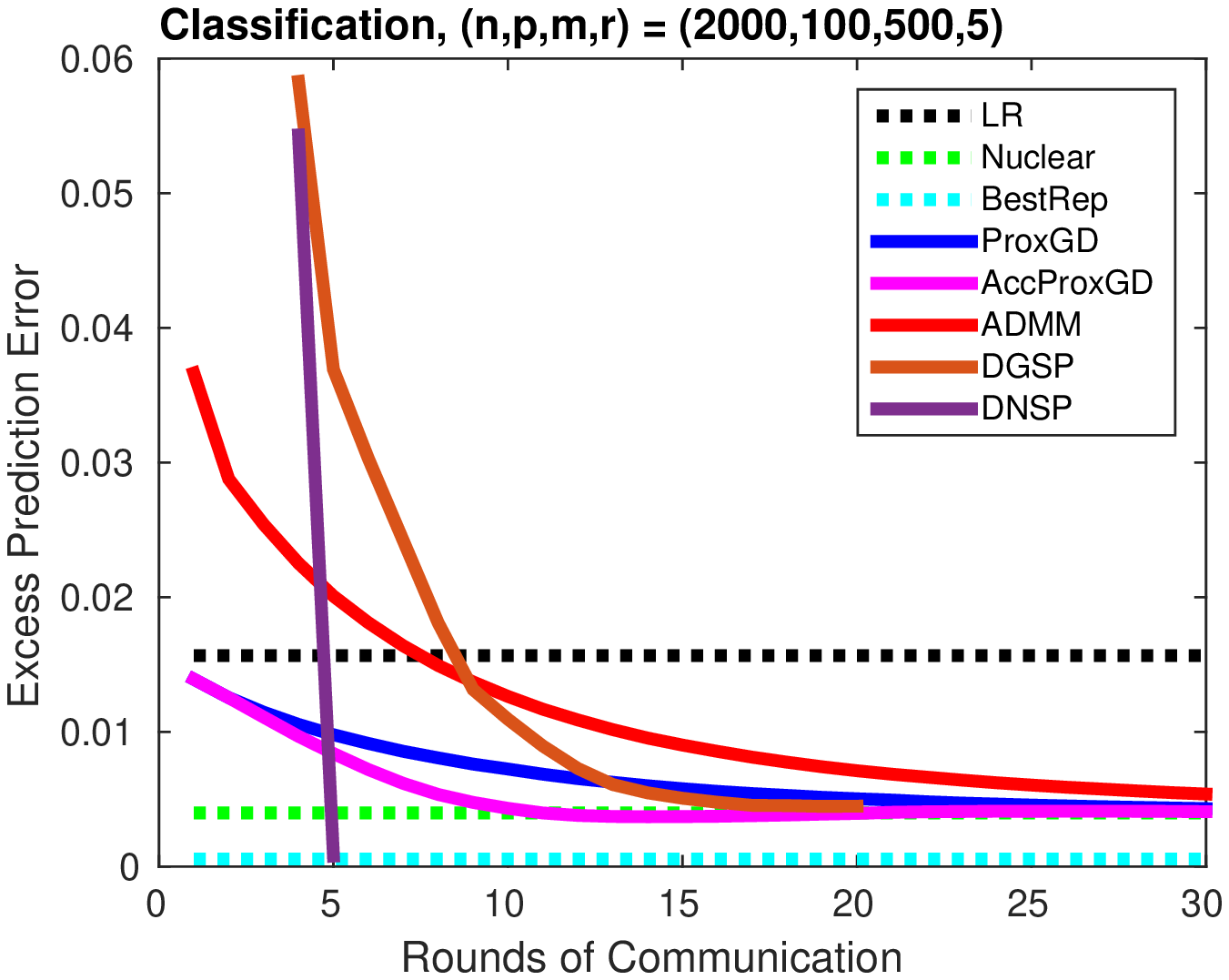}%
\includegraphics[width=0.33 \textwidth]{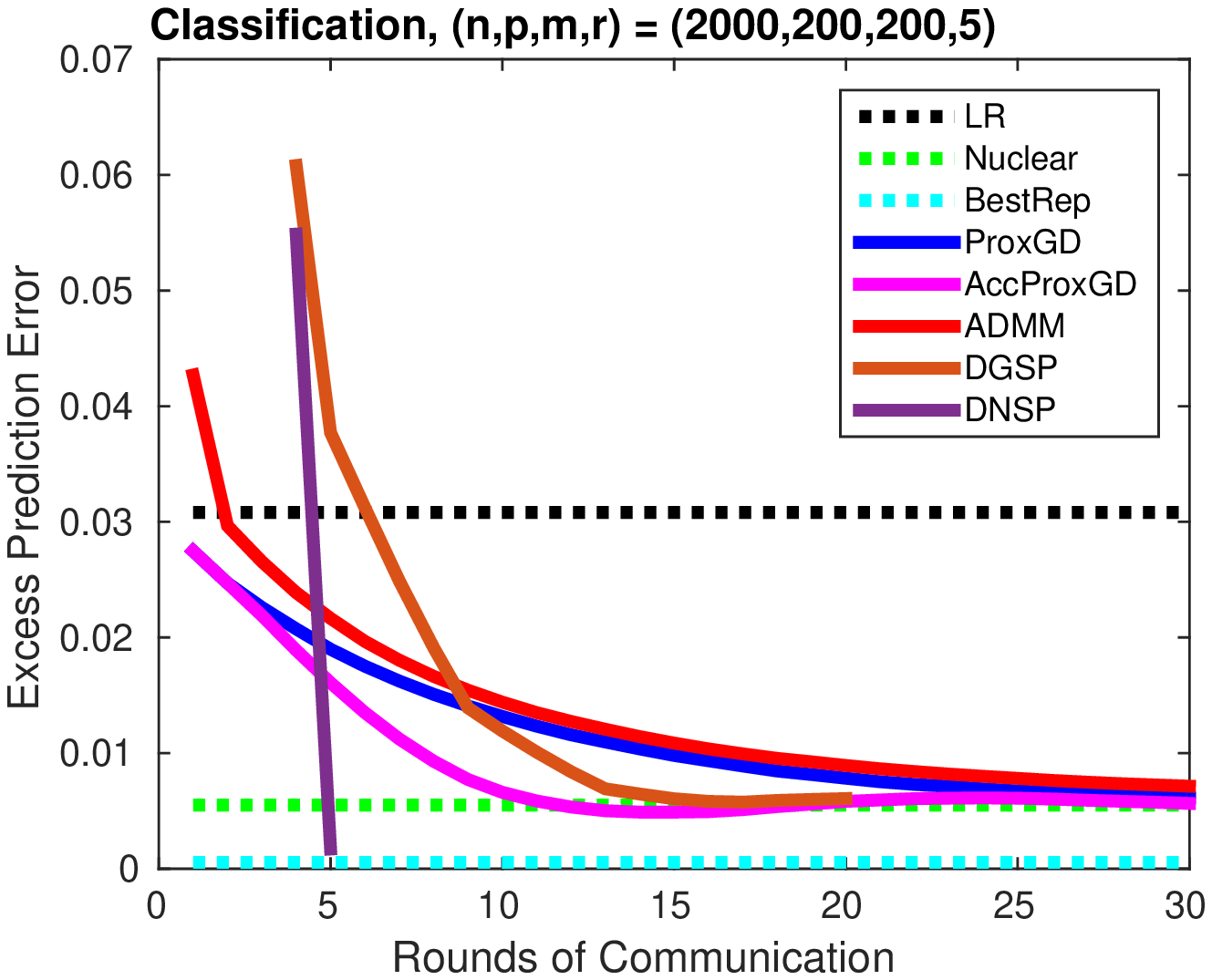}%
\end{center}
\caption{Excess prediction error for
  multi-task classification.}
\label{fig:simulation_classification}
\end{figure*}

\begin{figure*}[t]
\begin{center}
\includegraphics[width=0.33 \textwidth]{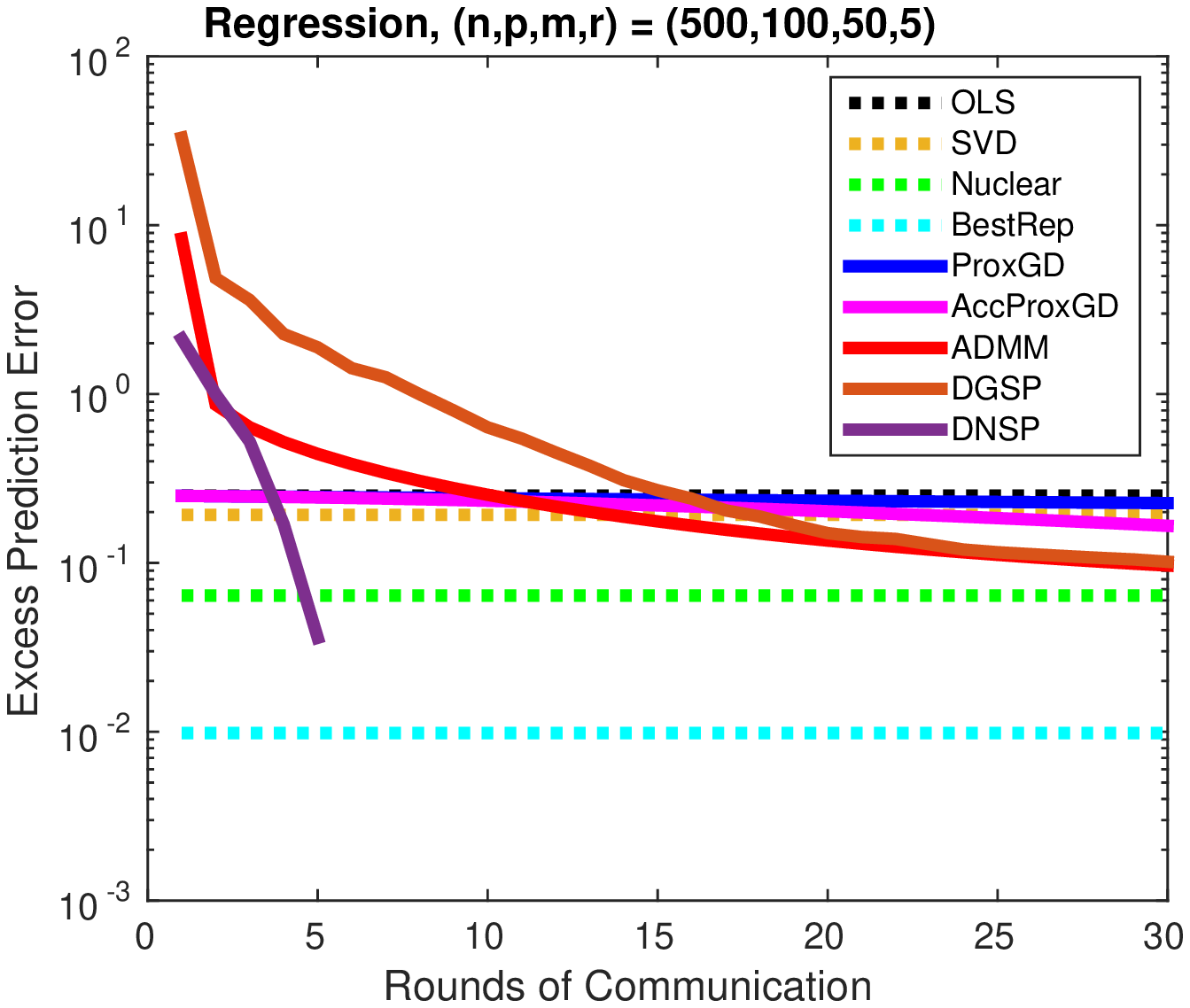}%
\includegraphics[width=0.33 \textwidth]{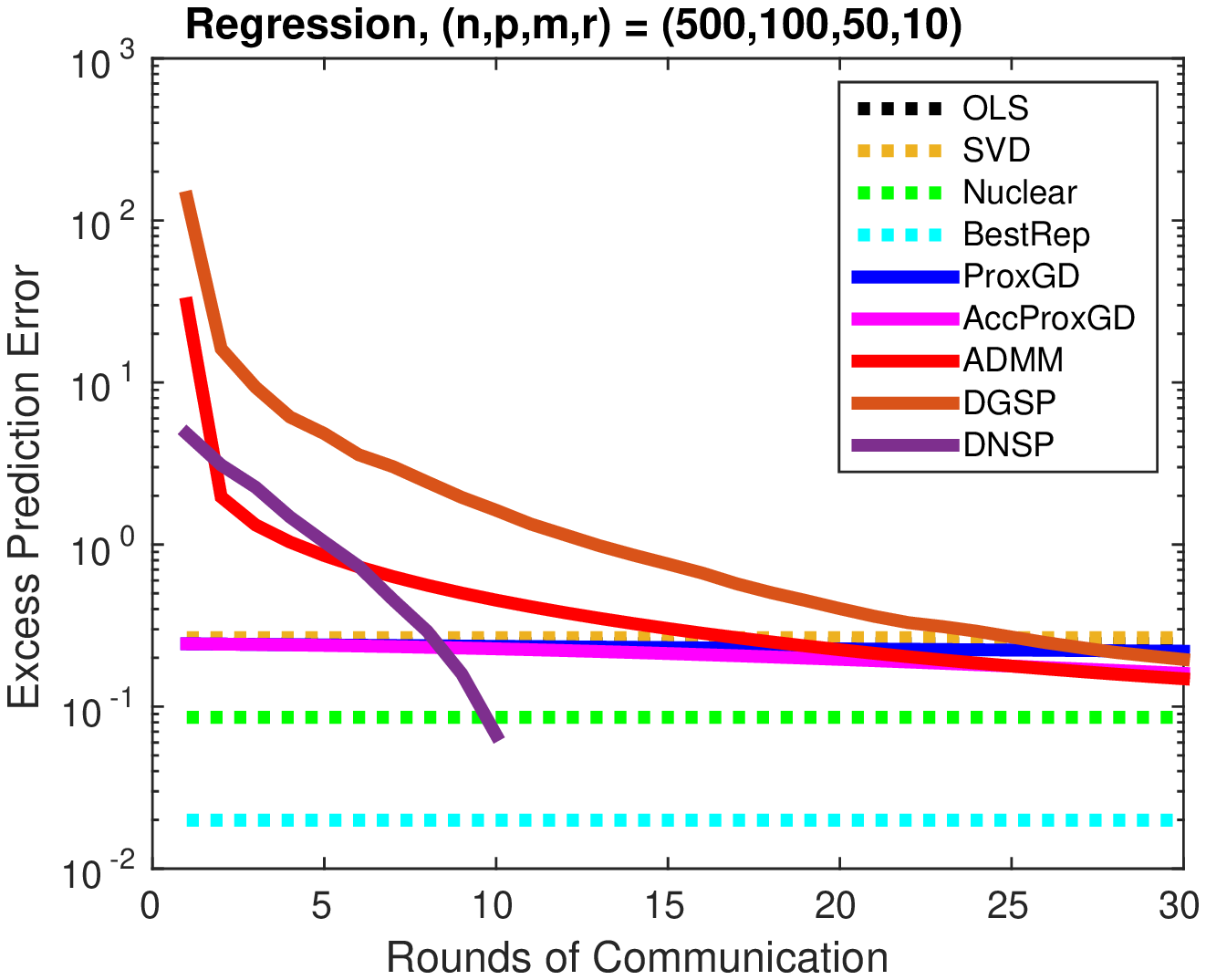}%
\includegraphics[width=0.33 \textwidth]{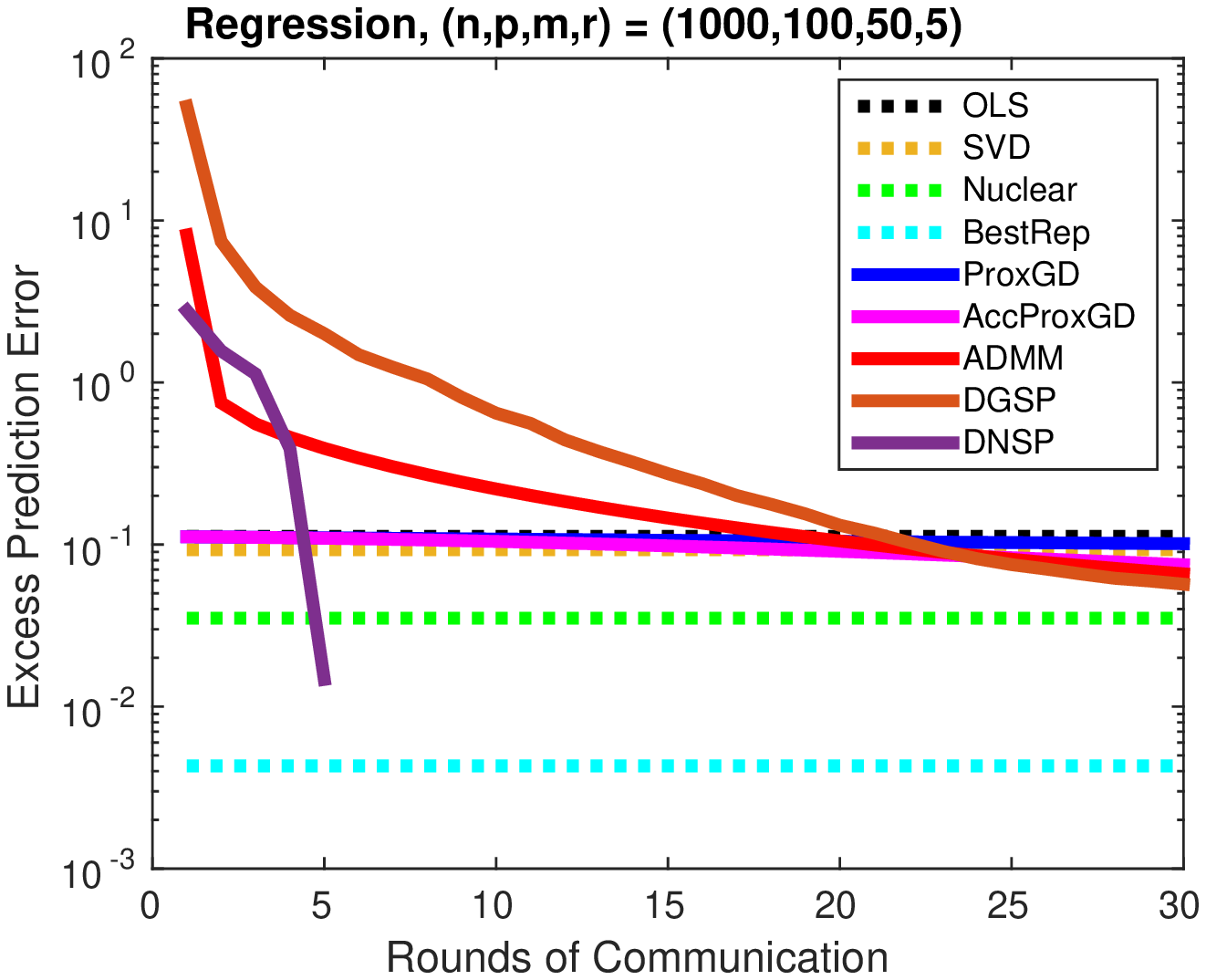}%
\end{center}
\caption{Excess prediction error for
  multi-task regression, with highly correlated features.}
\label{fig:simulation_regression_hard}
\end{figure*}

\vspace{-0.2 cm}
\section{Experiments}
\label{sec:experiments}
\vspace{-0.2 cm}

We first illustrate performance of different procedures on simulated data. We generate
data according to 
\begin{align*}
y_{ji} \mid \xb_{ji} \sim \Ncal(\wb_j^T\xb_{ji}, 1) 
\end{align*}
for regression problems and 
\begin{align*}
y_{ji} \mid \xb_{ji} \sim {\rm Bernoulli}\rbr{\rbr{1 + \exp(-\wb_j^T\xb_{ji})}^{-1}} 
\end{align*}
for classification problems. We generate the low-rank $W^*$ as
follows.  We first generate two matrices
$A \in \RR^{p \times r}, B \in \RR^{m \times r}$ with entries sampled
independently from a standard normal distribution. Then we extract the
left and right singular vectors of $AB^T$, denoted as
$U,V$. Finally, we set $W^* = USV^T$, where $S$ is a
diagonal matrix with exponentially decaying entries:
${\rm diag}(S) = [1,1/1.5,1/(1.5)^2, \ldots, 1/(1.5)^r]$.
 The feature
vectors $\xb_{ji}$ are sampled from a mean zero multivariate normal
with the covariance matrix $\Sigma = (\Sigma_{ab})_{a,b\in[p]}$,
$\Sigma_{ab} = 2^{-|a-b|}$. The regularization parameters for all
approaches were optimized to give the best prediction performance over
a held-out validation dataset. For \texttt{ProxGD} and \texttt{AccProxGD}, we initialized the solution from \texttt{Local}. Our simulation results are averaged
over 10 independent runs.
\removed{
\mcomment{We should probably include variance of the plots.  
Explain what do you generate anew between for each simulation run.
Do you generate one $W^*$ and then generate 10 different data-sets using the same 
parameter vector or $W^*$ also changes?}
}

We investigate how the performance of various procedures changes as a
function of problem parameters $(n,p,m,r)$. We compare the following
procedures: i) \texttt{Local}, where each machine solves an empirical
risk minimization problem (ordinary least squares or logistic
regression)
\removed{
\mcomment{Do we use any regularization here?}}.  ii)
Nuclear-norm regularization: which is a popular \texttt{Centralize}
approach: all machines send their data to the master, the master
solves a nuclear-norm regularized loss minimization problem. iii)
Learning with the best representation (\texttt{BestRep}): which
assumes the true projection matrix $U$ is known, and
just fit ordinal least squares or logistic regression model in the projected
low-dimensional subspace
\removed{
\mcomment{What do you mean by the best
  projection matrix here?}}. Note that this is not a practical
approach since in practice we do not know the best low-dimensional
representations of the data.  iv) Convex optimization approach which
runs distributed optimization algorithms over the nuclear
norm-regularized objective: here we implemented and compared the
following algorithms: distributed proximal gradient (\texttt{ProxGD});
distributed accelerated proximal gradient, (\texttt{AccProxGD});
distributed alternating direction method of multipliers
(\texttt{ADMM}); distributed Frank-Wolfe (\texttt{DFW})
\removed{  
  \mcomment{Explicitly write which
  figure the results are omitted from.}}.  v) The proposed
\texttt{DGSP} and \texttt{DNSP} approaches.  The simulation results
for regression and classification problems are shown in Figure
\ref{fig:simulation_regression} and
\ref{fig:simulation_classification}\footnote{For better visualization, here we
  omit the plot for \texttt{DFW} as its performance is
  significantly worse than others.}, respectively.  We plot how the
excess prediction error decreases as the number of rounds of
communications increases (\texttt{Local}, \texttt{Centralize} and
\texttt{BestRep} are one shot approaches thus the lines are
horizontal). From the plots, we have the following observations:
\begin{itemize}
\vspace{-0.4 cm}
\itemsep0em 
\item Nuclear norm regularization boosts the prediction performance
  over plain single task learning significantly, which shows clear
  advantage of leveraging the shared representation in multi-task
  learning.

\item \texttt{ADMM} and \texttt{AccProxGD} perform reasonably well
  \removed{  
  \mcomment{What does ``quite well'' mean here? In comparison to what?}},
  especially \texttt{ADMM}. One reason for the effectiveness of
  \texttt{ADMM} is that for the problem of nuclear norm regularized
  multi-task learning considered here, the \texttt{ADMM} update solves regularized ERM problems at every iteration. \texttt{ADMM} and \texttt{AccProxGD} clearly
  outperform \texttt{ProxGD}.

\item \texttt{ProxGD} and \texttt{DGSP} perform similarly.
  \texttt{DGSP} usually becomes worse as the iterations
  increases 
  \removed{
  \mcomment{Do you mean true rank of $W^*$ or rank of
    iterates?}}, while \texttt{ProxGD} converges to a global optimum
  of the nuclear norm regularized objective.

\item \texttt{DNSP} is the most communication-efficient method, and
  usually converges to a solution that is slightly better compared to
  the optimum of the nuclear regularization. This shows that
  second-order information helps a lot in reducing the communication
  cost.

\item The \texttt{DFW} performs the worst in most cases, even though
  \texttt{DFW} shares some similarity with \texttt{DGSP} in learning
  the subspace. The empirical results suggest the re-fitting
  step in \texttt{DGSP} is very important.
\end{itemize} 

\paragraph{One-shot SVD truncation} A natural question to ask is whether there exists a one-shot
communication method for the shared representation problem considered
here, that still matches the performance of centralized methods. One
reasonable solution is to consider the following SVD truncation
approach, which is based on the following derivation: consider the following well specified linear regression model:
\[
\yb_{ji} = \dotp{\xb_{ji}}{\wb_{j}^*} + \epsilon_{ji},
\]
where $\epsilon_{ji}$ is drawn from mean-zero Gaussian noise. It is easy to verify the following equation for OLS estimation:
\begin{align*}
\hat \wb_{{\rm local}(j)} = \wb^*_j + \rbr{\sum_{i} \xb_{ji} \xb_{ji}^T}^{-1} \rbr{\sum_{i} \epsilon_{ji} \xb_{ji}}.
\end{align*}
\removed{
\begin{align*}
\hat \wb_{{\rm local}(j)} =& \rbr{\sum_{i} \xb_{ji} \xb_{ji}^T}^{-1} \rbr{\sum_{i} \yb_{ji} \xb_{ji}} \\
=& \wb^*_j + \rbr{\sum_{i} \xb_{ji} \xb_{ji}^T}^{-1} \rbr{\sum_{i} \epsilon_{ji} \xb_{ji}}.
\end{align*}
}
Since ${\hat W_{\rm local}}$ is just $W^*$ plus some mean-zero Gaussian noise, it is natural to consider the following low-rank matrix denoising estimator:
\[
\min_{W} \norm{\hat W_{\rm local} - W}_F^2 \quad \text{\rm s.t.} \quad {\rm rank}(W) = r.
\]
where the solution is a simple SVD truncation, and can be implemented in a one-shot way: each worker send its \texttt{Local} solution to the master,
which then performs an SVD truncation step to maintain the top-$r$
components
\[
\hat W_{\rm svd} = U_r S_r V_r^T, \quad {\rm where} \quad USV^T = \texttt{SVD}({\hat W_{\rm local}}),
\]
and send the resulting estimation back to each worker, where $U_r,S_r,V_r$ are top-$r$ components of $U,S,V$. Though this approach might work well for some simple scenarios, but will generally fail when the features are highly correlated: 
although the \texttt{Local} solution $\hat W_{\rm local}$ can
output normal estimation of $W^*$, the estimation noise $\rbr{\sum_{i} \xb_{ji} \xb_{ji}^T}^{-1} \rbr{\sum_{i} \epsilon_{ji} \xb_{ji}}$ might be highly correlated
(depend on the correlation between features), which makes the SVD
truncation estimation not reliable. To illustrate this, consider a more complex
simulation which follows the same setup as above setting, except that
now the feature vectors $\xb_{ji}$ are sampled from a higher
correlation matrix $\Sigma = (\Sigma_{ab})_{a,b\in[p]}$,
$\Sigma_{ab} = 2^{-0.1|a-b|}$. The regression simulation results are
shown in Figure \ref{fig:simulation_regression_hard}, where we see
that the one-shot SVD truncation approach does not significantly
outperforms \texttt{Local}, sometimes even slightly worse.
\removed{\mcomment{Are there potentially other one-shot approaches that might work well? We 
only illustrate that one one-shot approach does not work.}}

Besides simulation, we also conducted extensive experiments on real world datasets, which are presented in Appendix \ref{sec:realworld} due to space limitation.

\vspace{-0.2 cm}
\section{Conclusion}
\label{sec:conclusion}
\vspace{-0.2 cm}

We studied the problem of distributed representation learning for
multiple tasks, discussed the implementation and guarantees for distributed convex optimization methods, and presented two
novel algorithms to learn low-dimensional projection in a greedy
way, which can be communication more efficient than
distributed convex optimization approaches. All approaches are
extensively evaluated on simulation and real world datasets.

\removed{
\section*{Acknowledgments}

This work is partially supported by an IBM Corporation Faculty
Research Fund at the University of Chicago Booth School of Business.
This work was completed in part with resources provided by the
University of Chicago Research Computing Center.
}

\clearpage
{
\bibliographystyle{my-plainnat}
\bibpunct{(}{)}{,}{a}{,}{,}
\bibliography{paper}

\begin{thebibliography}{49}
\providecommand{\natexlab}[1]{#1}
\providecommand{\url}[1]{\texttt{#1}}
\expandafter\ifx\csname urlstyle\endcsname\relax
  \providecommand{\doi}[1]{doi: #1}\else
  \providecommand{\doi}{doi: \begingroup \urlstyle{rm}\Url}\fi

\bibitem[Agarwal et~al.(2012)Agarwal, Negahban, and
  Wainwright]{Agarwal2012Fast}
A.~Agarwal, S.~Negahban, and M.~J. Wainwright.
\newblock Fast global convergence of gradient methods for high-dimensional
  statistical recovery.
\newblock \emph{Ann. Stat.}, 40\penalty0 (5):\penalty0 2452--2482, 2012.

\bibitem[Amit et~al.(2007)Amit, Fink, Srebro, and Ullman]{Amit2007Uncovering}
Y.~Amit, M.~Fink, N.~Srebro, and S.~Ullman.
\newblock Uncovering shared structures in multiclass classification.
\newblock In \emph{ICML}, pages 17--24. ACM, 2007.

\bibitem[Ando and Zhang(2005)]{Ando2005framework}
R.~K. Ando and T.~Zhang.
\newblock A framework for learning predictive structures from multiple tasks
  and unlabeled data.
\newblock \emph{J. Mach. Learn. Res.}, 6:\penalty0 1817--1853, 2005.

\bibitem[{Argyriou} et~al.(2008){Argyriou}, {Evgeniou}, and
  {Pontil}]{argyriou08convex}
A.~{Argyriou}, T.~{Evgeniou}, and M.~{Pontil}.
\newblock Convex multi-task feature learning.
\newblock \emph{Mach. Learn.}, 73\penalty0 (3):\penalty0 243--272, 2008.

\bibitem[Bach et~al.(2011)Bach, Jenatton, Mairal, and
  Obozinski]{Bach2011Optimization}
F.~Bach, R.~Jenatton, J.~Mairal, and G.~Obozinski.
\newblock Optimization with sparsity-inducing penalties.
\newblock \emph{Found. Trends Mach. Learn.}, 4\penalty0 (1):\penalty0 1--106,
  2011.

\bibitem[Balcan et~al.(2012)Balcan, Blum, Fine, and
  Mansour]{Balcan2012Distributed}
M.-F. Balcan, A.~Blum, S.~Fine, and Y.~Mansour.
\newblock Distributed learning, communication complexity and privacy.
\newblock In \emph{JMLR W\&CP 23: COLT 2012}, volume~23, pages 26.1--26.22,
  2012.

\bibitem[Bartlett and Mendelson(2002)]{bartlett2003rademacher}
P.~L. Bartlett and S.~Mendelson.
\newblock Rademacher and gaussian complexities: Risk bounds and structural
  results.
\newblock \emph{J. Mach. Learn. Res.}, 3:\penalty0 463--482, 2002.

\bibitem[Bekkerman et~al.(2011)Bekkerman, Bilenko, and
  Langford]{Bekkerman2011Scaling}
R.~Bekkerman, M.~Bilenko, and J.~Langford.
\newblock \emph{Scaling up machine learning: Parallel and distributed
  approaches}.
\newblock Cambridge University Press, 2011.

\bibitem[Bellet et~al.(2015)Bellet, Liang, Garakani, Balcan, and
  Sha]{Bellet2015Distributed}
A.~Bellet, Y.~Liang, A.~B. Garakani, M.-F. Balcan, and F.~Sha.
\newblock A distributed frank-wolfe algorithm for communication-efficient
  sparse learning.
\newblock In \emph{SDM}, pages 478--486. 2015.

\bibitem[Bousquet and Bottou(2008)]{bousquet2008tradeoffs}
O.~Bousquet and L.~Bottou.
\newblock The tradeoffs of large scale learning.
\newblock In \emph{NIPS}, pages 161--168, 2008.

\bibitem[Boyd et~al.(2011)Boyd, Parikh, Chu, Peleato, and
  Eckstein]{Boyd2011Distributed}
S.~P. Boyd, N.~Parikh, E.~Chu, B.~Peleato, and J.~Eckstein.
\newblock Distributed optimization and statistical learning via the alternating
  direction method of multipliers.
\newblock \emph{Found. Trends Mach. Learn.}, 3\penalty0 (1):\penalty0 1--122,
  2011.

\bibitem[Cai et~al.(2010)Cai, Cand{\`e}s, and Shen]{cai2010singular}
J.-F. Cai, E.~J. Cand{\`e}s, and Z.~Shen.
\newblock A singular value thresholding algorithm for matrix completion.
\newblock \emph{SIAM Journal on Optimization}, 20\penalty0 (4):\penalty0
  1956--1982, 2010.

\bibitem[Caruana(1997)]{caruana1997multitask}
R.~Caruana.
\newblock Multitask learning.
\newblock \emph{Mach. Learn.}, 28\penalty0 (1):\penalty0 41--75, 1997.

\bibitem[Chapelle et~al.(2010)Chapelle, Shivaswamy, Vadrevu, Weinberger, Zhang,
  and Tseng]{chapelle2010multi}
O.~Chapelle, P.~Shivaswamy, S.~Vadrevu, K.~Weinberger, Y.~Zhang, and B.~Tseng.
\newblock Multi-task learning for boosting with application to web search
  ranking.
\newblock In \emph{KDD}, pages 1189--1198. ACM, 2010.

\bibitem[Collobert et~al.(2011)Collobert, Weston, Bottou, Karlen, Kavukcuoglu,
  and Kuksa]{collobert2011natural}
R.~Collobert, J.~Weston, L.~Bottou, M.~Karlen, K.~Kavukcuoglu, and P.~Kuksa.
\newblock Natural language processing (almost) from scratch.
\newblock \emph{J. Mach. Learn. Res.}, 12:\penalty0 2493--2537, 2011.

\bibitem[Frank and Wolfe(1956)]{frank1956algorithm}
M.~Frank and P.~Wolfe.
\newblock An algorithm for quadratic programming.
\newblock \emph{Naval research logistics quarterly}, 3\penalty0 (1-2):\penalty0
  95--110, 1956.

\bibitem[He and Yuan(2012)]{He2012Convergence}
B.~He and X.~Yuan.
\newblock On the $o(1/n)$ convergence rate of the douglas-rachford alternating
  direction method.
\newblock \emph{SIAM Journal on Numerical Analysis}, 50\penalty0 (2):\penalty0
  700--709, 2012.

\bibitem[Hong and Luo(2012)]{hong2012linear}
M.~Hong and Z.-Q. Luo.
\newblock On the linear convergence of the alternating direction method of
  multipliers.
\newblock \emph{ArXiv e-prints, arXiv:1208.3922}, 2012.

\bibitem[Jaggi(2013)]{jaggi2013revisiting}
M.~Jaggi.
\newblock Revisiting frank-wolfe: Projection-free sparse convex optimization.
\newblock In \emph{ICML}, pages 427--435, 2013.

\bibitem[Jaggi et~al.(2014)Jaggi, Smith, Tak{\'a}c, Terhorst, Krishnan,
  Hofmann, and Jordan]{Jaggi2014Communication}
M.~Jaggi, V.~Smith, M.~Tak{\'a}c, J.~Terhorst, S.~Krishnan, T.~Hofmann, and
  M.~I. Jordan.
\newblock Communication-efficient distributed dual coordinate ascent.
\newblock In \emph{NIPS}, pages 3068--3076, 2014.

\bibitem[Jain et~al.(2013)Jain, Netrapalli, and Sanghavi]{jain2013low}
P.~Jain, P.~Netrapalli, and S.~Sanghavi.
\newblock Low-rank matrix completion using alternating minimization.
\newblock In \emph{STOC}, pages 665--674. ACM, 2013.

\bibitem[Ji and Ye(2009)]{Ji2009accelerated}
S.~Ji and J.~Ye.
\newblock An accelerated gradient method for trace norm minimization.
\newblock In \emph{ICML}, pages 457--464. ACM, 2009.

\bibitem[Kim and Xing(2010)]{kim2010tree}
S.~Kim and E.~P. Xing.
\newblock Tree-guided group lasso for multi-task regression with structured
  sparsity.
\newblock In \emph{ICML}, pages 543--550, 2010.

\bibitem[Lacoste-Julien and Jaggi(2015)]{lacoste2015global}
S.~Lacoste-Julien and M.~Jaggi.
\newblock On the global linear convergence of frank-wolfe optimization
  variants.
\newblock In \emph{NIPS}, pages 496--504, 2015.

\bibitem[Lapin et~al.(2014)Lapin, Schiele, and Hein]{lapin2014scalable}
M.~Lapin, B.~Schiele, and M.~Hein.
\newblock Scalable multitask representation learning for scene classification.
\newblock In \emph{CVPR}, pages 1434--1441, 2014.

\bibitem[Lee et~al.(2015)Lee, Sun, Liu, and Taylor]{lee2015communication}
J.~D. Lee, Y.~Sun, Q.~Liu, and J.~E. Taylor.
\newblock Communication-efficient sparse regression: a one-shot approach.
\newblock \emph{ArXiv e-prints, arXiv:1503.04337}, 2015.

\bibitem[Lenk et~al.(1996)Lenk, DeSarbo, Green, and
  Young]{lenk1996hierarchical}
P.~J. Lenk, W.~S. DeSarbo, P.~E. Green, and M.~R. Young.
\newblock Hierarchical bayes conjoint analysis: Recovery of partworth
  heterogeneity from reduced experimental designs.
\newblock \emph{Marketing Science}, 15\penalty0 (2):\penalty0 173--191, 1996.

\bibitem[Maurer and Pontil(2013)]{Maurer2012Excess}
A.~Maurer and M.~Pontil.
\newblock Excess risk bounds for multitask learning with trace norm
  regularization.
\newblock pages 55--76, 2013.

\bibitem[Nesterov(1983)]{nesterov1983method}
Y.~Nesterov.
\newblock A method of solving a convex programming problem with convergence
  rate ${\cal o}(1/k^2)$.
\newblock In \emph{Soviet Mathematics Doklady}, volume~27, pages 372--376,
  1983.

\bibitem[Ram et~al.(2010)Ram, Nedi{\'c}, and Veeravalli]{Ram2010Distributed}
S.~S. Ram, A.~Nedi{\'c}, and V.~V. Veeravalli.
\newblock Distributed stochastic subgradient projection algorithms for convex
  optimization.
\newblock \emph{Journal of optimization theory and applications}, 147\penalty0
  (3):\penalty0 516--545, 2010.

\bibitem[Sander and Schneider(1991)]{Sander1991Database}
C.~Sander and R.~Schneider.
\newblock Database of homology-derived protein structures and the structural
  meaning of sequence alignment.
\newblock \emph{Proteins: Structure, Function, and Bioinformatics}, pages
  56--68, 1991.

\bibitem[Seltzer and Droppo(2013)]{seltzer2013multi}
M.~L. Seltzer and J.~Droppo.
\newblock Multi-task learning in deep neural networks for improved phoneme
  recognition.
\newblock In \emph{ICASSP}, pages 6965--6969. IEEE, 2013.

\bibitem[Shalev-Shwartz and Ben-David(2014)]{Shalev-Shwartz2014Understanding}
S.~Shalev-Shwartz and S.~Ben-David.
\newblock \emph{Understanding machine learning: From theory to algorithms}.
\newblock Cambridge University Press, 2014.

\bibitem[Shalev-Shwartz et~al.(2010)Shalev-Shwartz, Srebro, and
  Zhang]{Shalev-Shwartz2010Trading}
S.~Shalev-Shwartz, N.~Srebro, and T.~Zhang.
\newblock Trading accuracy for sparsity in optimization problems with sparsity
  constraints.
\newblock \emph{SIAM Journal on Optimization}, 20\penalty0 (6):\penalty0
  2807--2832, 2010.

\bibitem[Shalev-Shwartz et~al.(2011)Shalev-Shwartz, Gonen, and
  Shamir]{Shalev-Shwartz2011Large}
S.~Shalev-Shwartz, A.~Gonen, and O.~Shamir.
\newblock Large-scale convex minimization with a low-rank constraint.
\newblock In \emph{ICML}, 2011.

\bibitem[Shamir and Srebro(2014)]{Shamir2014Distributed}
O.~Shamir and N.~Srebro.
\newblock Distributed stochastic optimization and learning.
\newblock In \emph{Allerton}, pages 850--857. IEEE, 2014.

\bibitem[Shamir et~al.(2014)Shamir, Srebro, and Zhang]{Shamir2013Communication}
O.~Shamir, N.~Srebro, and T.~Zhang.
\newblock Communication efficient distributed optimization using an approximate
  newton-type method.
\newblock In \emph{ICML}, pages 1000--1008, 2014.

\bibitem[Spyromitros-Xioufis et~al.(2012)Spyromitros-Xioufis, Tsoumakas,
  Groves, and Vlahavas]{Spyromitros-Xioufis2012Multi}
E.~Spyromitros-Xioufis, G.~Tsoumakas, W.~Groves, and I.~Vlahavas.
\newblock Multi-target regression via input space expansion: Treating targets
  as inputs.
\newblock \emph{ArXiv e-prints, arXiv:1211.6581}, 2012.

\bibitem[Tropp(2004)]{tropp2004greed}
J.~A. Tropp.
\newblock Greed is good: Algorithmic results for sparse approximation.
\newblock \emph{IEEEit}, 50\penalty0 (10):\penalty0 2231--2242, 2004.

\bibitem[Turnbull et~al.(2008)Turnbull, Barrington, Torres, and
  Lanckriet]{turnbull2008semantic}
D.~Turnbull, L.~Barrington, D.~Torres, and G.~Lanckriet.
\newblock Semantic annotation and retrieval of music and sound effects.
\newblock \emph{IEEE Transactions on Acoustics, Speech and Signal Processing},
  16\penalty0 (2):\penalty0 467--476, 2008.

\bibitem[Wang et~al.(2015{\natexlab{a}})Wang, Kolar, and
  Srebro]{Wang2015Distributed}
J.~Wang, M.~Kolar, and N.~Srebro.
\newblock Distributed multitask learning.
\newblock \emph{ArXiv e-prints, arXiv:1510.00633}, 2015{\natexlab{a}}.

\bibitem[Wang et~al.(2015{\natexlab{b}})Wang, Lai, Lu, Fan, Davulcu, and
  Ye]{wang2015orthogonal}
Z.~Wang, M.-J. Lai, Z.~Lu, W.~Fan, H.~Davulcu, and J.~Ye.
\newblock Orthogonal rank-one matrix pursuit for low rank matrix completion.
\newblock \emph{SIAM Journal on Scientific Computing}, 37\penalty0
  (1):\penalty0 A488--A514, 2015{\natexlab{b}}.

\bibitem[Weinberger et~al.(2009)Weinberger, Dasgupta, Langford, Smola, and
  Attenberg]{weinberger2009feature}
K.~Weinberger, A.~Dasgupta, J.~Langford, A.~Smola, and J.~Attenberg.
\newblock Feature hashing for large scale multitask learning.
\newblock In \emph{ICML}, pages 1113--1120. ACM, 2009.

\bibitem[Xue et~al.(2007)Xue, Liao, Carin, and Krishnapuram]{xue2007multi}
Y.~Xue, X.~Liao, L.~Carin, and B.~Krishnapuram.
\newblock Multi-task learning for classification with dirichlet process priors.
\newblock \emph{J. Mach. Learn. Res.}, 8:\penalty0 35--63, 2007.

\bibitem[Yuan et~al.(2007)Yuan, Ekici, Lu, and Monteiro]{yuan2007dimension}
M.~Yuan, A.~Ekici, Z.~Lu, and R.~Monteiro.
\newblock Dimension reduction and coefficient estimation in multivariate linear
  regression.
\newblock \emph{J. R. Stat. Soc. B}, 69\penalty0 (3):\penalty0 329--346, 2007.

\bibitem[Zhang and Xiao(2015)]{zhang2015communication}
Y.~Zhang and L.~Xiao.
\newblock Communication-efficient distributed optimization of self-concordant
  empirical loss.
\newblock \emph{ArXiv e-prints, arXiv:1501.00263}, 2015.

\bibitem[Zhang et~al.(2012)Zhang, Wainwright, and
  Duchi]{Zhang2012Communication}
Y.~Zhang, M.~J. Wainwright, and J.~C. Duchi.
\newblock Communication-efficient algorithms for statistical optimization.
\newblock In \emph{NIPS}, pages 1502--1510, 2012.

\bibitem[Zhang et~al.(2013)Zhang, Duchi, Jordan, and
  Wainwright]{Zhang2013Information}
Y.~Zhang, J.~C. Duchi, M.~I. Jordan, and M.~J. Wainwright.
\newblock Information-theoretic lower bounds for distributed statistical
  estimation with communication constraints.
\newblock In \emph{NIPS}, pages 2328--2336, 2013.

\bibitem[Zhou et~al.(2013)Zhou, Liu, Narayan, and Ye]{zhou2013modeling}
J.~Zhou, J.~Liu, V.~A. Narayan, and J.~Ye.
\newblock Modeling disease progression via multi-task learning.
\newblock \emph{NeuroImage}, 78:\penalty0 233 -- 248, 2013.

\end{thebibliography}
}

\clearpage
\section*{Appendix}
\appendix
\section{Distributed Alternating Direction Methods of Multipliers}
\label{sec:admm}

\begin{algorithm}[p]
\SetAlgoLined
\For{$t=1, 2, \ldots $}{
\underline{\textbf{Workers:}}\\
\For{$j=1, 2, \ldots, m$}{
Each worker solves the regularized ERM problem as \eqref{eq:admm_update_w} to get $\wb_j^{(t+1)}$, and send it to the master; \\
\texttt{Wait}; \\
Receive $\zb_j^{(t+1)},\qb_j^{(t+1)}$ from master.
}
\underline{\textbf{Master:}}\\
\If{Receive $\wb_j^{(t+1)}$ from all workers}{
Concatenate the current solutions $\wb_j^{(t+1)}$, and update $Z^{(t+1)}$ as \eqref{eq:admm_update_z}; \\
Update $Q^{(t+1)}$ as \eqref{eq:admm_update_q}; \\
Send $\zb_j^{(t+1)},\qb_j^{(t+1)}$ to the corresponding worker.
}
}
\caption{\texttt{ADMM}: Distributed ADMM for Multi-Task Learning.}
\label{alg:admm}
\end{algorithm}

The Alternating Direction Methods of Multipliers (ADMM) is also a
popular method for distributed optimization
\citep{Boyd2011Distributed} and can be used to solve the distributed
low-rank multi-task learning problem. We first write the objective \eqref{eq:nnmtl} as
\begin{align*}
\arg\min_{W,Z} ~ & \Lcal_n(W) + \lambda \norm{Z}_*, \quad {\rm subject\ to} \quad W = Z.
\end{align*}
By introducing the Lagrangian and augmented terms, 
we get the following unconstrained problem:
\begin{align*}
\tilde \Lcal(W,Z,Q) =& 
\Lcal_n(W) + \lambda \norm{Z}_* + \dotp{W-Z}{Q} \nonumber \\
&+ \frac{\rho}{2} \norm{W-Z}_F^2,
\end{align*}
where $\rho$ is a parameter controlling the augmentation level. Note
that except for $Z$, the augmented Lagrangian objective are
decomposable across tasks. To implement the distributed ADMM
algorithm, we let the workers  maintain the data and $W$, while 
the master maintains $Z$ and $Q$.  
At round $t$, each machine separately solves
\begin{align}
\wb_j^{(t+1)} = \arg\min_{\wb} &\Lcal_{nj}(\wb_j)
 + \dotp{\wb_j^{(t+1)} - \zb_j^{(t)}}{\qb_j^{(t)}} \nonumber \\
 &+ \frac{\rho}{2} \norm{\wb_j^{(t+1)} - \zb_j^{(t)}}_2, 
\label{eq:admm_update_w}
\end{align}
which is minimizing the local loss plus a regularization term.  Next,
each worker sends their solution to the master, which performs
the following updates for $Z$ and $Q$
\begin{align}
Z^{(t+1)} =& \arg\min_{Z}~\dotp{W^{(t+1)} - Z}{Q^{t}}  + \lambda \norm{Z}_* \nonumber \\
&+ \frac{\rho}{2} \norm{W^{(t+1)} - Z}_F^2,  \label{eq:admm_update_z} \\
Q^{(t+1)} =& Q^{(t)} + \rho(W^{(t+1)} - Z^{(t+1)}), \label{eq:admm_update_q}
\end{align}
which have closed-form solutions.

The algorithm \texttt{ADMM} is summarized in Algorithm
\ref{alg:admm}. Note that compared to methods discussed before,
\texttt{ADMM} needs to communicate three $p$-dimensional vectors
between each worker and the master at each round, while the proximal
gradient approaches only communicate two $p$-dimensional vectors per
round. Based on convergence results of ADMM \citep{He2012Convergence},
$\Ocal\rbr{\frac{mA^2}{\varepsilon}}$ rounds of communication are
needed to obtain $\varepsilon$-generalization error.

\section{Distributed Frank-Wolfe Method}
\label{sec:dfw}

\begin{algorithm}[t]
\SetAlgoLined
\For{$t=0, 2, \ldots $}{
\underline{\textbf{Workers:}}\\
\For{$j=1, 2, \ldots, m$}{
Each worker compute the its gradient direction $ \nabla \Lcal_{nj}(\wb_j^{(t)}) $, and send it to the master; \\
}
\If{Receive $\vb_j \ub$ from the master}{
Set $\gamma = \frac{2}{t+2}$;\\
Update $\wb^{(t+1)}_j$ as \eqref{eq:dfw}.
}
\underline{\textbf{Master:}}\\
\If{Receive $\nabla \Lcal_{nj}(\wb_j^{(t)}) $ from all workers}{
Concatenate the gradient vectors, and compute the largest singular vectors: $(\ub,\vb) = \textsf{SV}(\nabla \Lcal_n(W^{(t)}))$;\\
Send $\vb_j \ub$ to $j$-th worker.
}
}
\caption{\texttt{DFW}: Distributed Frank-Wolfe for Multi-Task Learning.}
\label{alg:dfw}
\end{algorithm}

Another approach we consider is the distributed Frank-Wolfe method
\citep{frank1956algorithm,jaggi2013revisiting,Bellet2015Distributed}.
This methods does not require performing SVD, which might bring
additional computational advantages. Instead of directly minimizing the
nuclear norm regularized objective, the Frank-Wolfe algorithm considers
the equivalent constrained minimization problem
\begin{align*}
\min_{W} \Lcal_n(W) \quad {\rm subject\ to} \quad \norm{W}_* \leq R.
\end{align*}
At each step, Frank-Wolfe algorithm considers the following direction to update
\begin{align*}
Z^{(t)} = \arg\min_{\norm{Z}_* \leq R} \dotp{\nabla \Lcal_n(W^{(t)})}{Z} = 
-R \cdot \ub \vb^T,
\end{align*}
where $(\ub,\vb) = \textsf{SV}(\nabla \Lcal_n(W^{(t)}))$ is the
leading singular vectors of $\nabla \Lcal_n(W^{(t)})$. The next iterate is obtained as
\begin{align*}
W^{(t+1)} = (1-\gamma)W^{(t)} + \gamma Z^{(t)},
\end{align*}
where $\gamma$ is a step size parameter.  To implement this algorithm
in a distributed way, the master first collects the gradient matrix
$\nabla \Lcal_n(W^{(t)})$ and computes $\ub$ and $\vb$. The
vector $\vb_j \ub$ is sent to $j$-th machine, which performs the
following update:
\begin{align}
\wb_j^{(t+1)} = (1-\gamma)\wb_j^{(t)} - \gamma R \vb_j \ub.
\label{eq:dfw}
\end{align} 

The algorithm is summarized in Algorithm \ref{alg:dfw}.  Similar to
the distributed (accelerated) proximal gradient descent, the
distributed Frank-Wolfe only requires communication of two
$p$-dimensional vectors per round. Though computationally cheaper
compared to other methods considered in this section, the distributed
Frank-Wolfe algorithm enjoys similar convergence guarantees to the
distributed proximal gradient descent
\citep{jaggi2013revisiting}, that is, after
$\Ocal\rbr{\frac{mHA^2}{\varepsilon}}$ iterations, the solution will be $\varepsilon$ suboptimal.

\section{Pseudocode of the algorithms}

\begin{algorithm}[t]
\SetAlgoLined
\For{$t=1, 2, \ldots $}{
\underline{\textbf{Workers:}}\\
\For{$j=1, 2, \ldots, m$}{
Each worker compute the its gradient direction $ \nabla  \Lcal_{nj}(\wb_j^{(t)})= \frac{1}{mn} \sum_{i=1}^n \ell'(\dotp{\wb^{(t)}_j}{\xb_{ji}}, y_{ji}) \xb_{ji}$, and send it to the master; \\
\texttt{Wait}; \\
Receive $\wb_j^{(t+1)}$ from master.
}
\underline{\textbf{Master:}}\\
\If{Receive $\nabla \Lcal_{nj}(\wb_j^{(t)})$ from all workers}{
Concatenate the gradient vectors, and update $W^{(t+1)}$ as \eqref{eq:prox_op}; \\
Send $\wb_j^{(t+1)}$ to all workers.
}
}
\caption{\texttt{ProxGD}: Distributed Proximal Gradient.}
\label{alg:prox_gd}
\end{algorithm}

\begin{algorithm}[t]
\SetAlgoLined
\For{$t=1, 2, \ldots $}{
\underline{\textbf{Workers:}}\\
\For{$j=1, 2, \ldots, m$}{
Each worker compute the its gradient direction $ \nabla \Lcal_n(\zb_j^{(t)})= \frac{1}{mn} \sum_{i=1}^n \ell'(\dotp{\zb^{(t)}_j}{\xb_{ji}}, y_{ji}) \xb_{ji}$, and send it to the master; \\
\texttt{Wait}; \\
Receive $\zb_j^{(t+1)}$ from master.
}
\underline{\textbf{Master:}}\\
\If{Receive $\nabla \Lcal_n(\zb_j^{(t)})$ from all workers}{
Concatenate the gradient vectors, and update $W^{(t+1)}$ as \eqref{eq:prox_op_z}; \\
Update $Z^{(t+1)}$ as \eqref{eq:acc_prox}; \\
Send $\zb_j^{(t+1)}$ to all workers.
}
}
\caption{\texttt{AccProxGD}: Accelerated Distributed Proximal Gradient for Multi-Task Learning.}
\label{alg:acc_prox_gd}
\end{algorithm}

\begin{algorithm}[t]
\SetAlgoLined
\For{$t=1, 2, \ldots$}{
\underline{\textbf{Workers:}}\\
\For{$j=1, 2, \ldots, m$}{
Each worker computes the Newton direction $\Delta \Lcal_{nj}(\wb_t^{(t)}) = 
\rbr{\nabla^2 \Lcal_{nj}(\wb_t^{(t)})}^{-1} \nabla \Lcal_{nj}(\wb_t^{(t)})$ and sends it to the master.
}
\If{Receive $\ub$ from the master}{
Perform Gram-Schmidt orthogonalization: \\
$\ub \leftarrow \ub - \sum_{k=1}^{t-1} \dotp{U_k}{\ub}$; \\
Normalize $\ub = \ub/\norm{\ub}_2$;\\
Update the projection matrix $U = [U ~ \ub]$;\\
Solve the projected ERM problem: \\
$\vb_j = \arg\min_{\vb_j} \frac{1}{n} \sum_{i=1}^n \ell(\dotp{\vb_j}{U^T X_{ji}},y_{ji})$;\\
Update $\wb^{(t+1)}_j = U \vb_j$.
}
\underline{\textbf{Master:}}\\
\If{Receive $\Delta \Lcal_{nj}(\wb_t^{(t)})$ from all workers}{
Concatenate the Newton vectors, and compute the largest singular vectors: $(\ub,\vb) = \textsf{SV}(\Delta \Lcal_n(W^{(t)}))$;\\
Send $\ub$ to all workers.
}
}
\caption{\texttt{DNSP}: Distributed Newton Subspace Pursuit.}
\label{alg:dnsp}
\end{algorithm}

\section{Proof of Proposition \ref{prop:orthogonal}}
\label{sec:orthogonal}
\begin{proof}
  It is sufficient to prove that at every iteration, the current
  projection matrix $U$ and the subspace to be added $\ub$ are
  orthogonal to each other. Note that by the optimality condition:
\begin{align*}
\nabla_V \rbr{\Lcal_n(UV^T)} =  U^T \nabla \Lcal_n(W^{(t)}) = 0.
\end{align*}
Since $\ub$ is the leading left singular vector of
$\nabla \Lcal_n(W^{(t)})$, we have $U^T \ub = 0$.  Each column of $U$
has unit length, since it is a left singular vector of some matrix.
\end{proof}

\section{Proof of Proposition \ref{prop:smoothness}}
\label{sec:proofprop}
\begin{proof}
  It is sufficient to prove that the largest eigenvalue of
  $\nabla^2 \Lcal_n(W)$ does not exceed $H$. Since
  $\nabla^2 \Lcal_n(W)$ is a block diagonal matrix, it is sufficient
  to show that for every block $j \in [m]$, the largest eigenvalue of
  the block $\nabla^2 \Lcal_{nj}(\wb_j)$ is not larger than $H$. 
  \removed{ 
  \mcomment{Add details here. This is not clear.}
  
  {\bf Todo: This is still not clear.}  We have 
  \[
    \nabla^2 \Lcal_{nj}(\wb_j) 
    = n^{-1}\sum_{i=1}^n \nabla^2 \ell(\wb_j^T\xb_{ji}, y_{ji})
    = n^{-1}\sum_{i=1}^n \ell''(\wb_j^T\xb_{ji}, y_{ji}) \xb_{ji}\xb_{ji}^T.
  \]
  How do you use $H$-smoothness in Assumption~\ref{assum:smoothness} to say that 
  $|\ell''(\wb_j^T\xb_{ji}, y_{ji})| \leq H$?
}

  This is
  true by the $H$-smoothness of $\ell(\cdot)$ and the fact that the
  data points have bounded length:
  \[
  \norm{\nabla^2 \Lcal_{nj}(\wb_j)}_2 \leq H \cdot \max_{i,j} \norm{\xb_{ji}}_2 \leq H. 
  \]
\end{proof}

\section{Proof of Theorem \ref{thm:dgsp}}
\label{sec:proofthm}

\begin{proof}
By the smoothness of $\Lcal_n$, we know
\begin{align}
\Lcal_n(W^{(t+1)}) \leq& \min_{b} \Lcal_n(W^{(t)} + b \ub \vb^T)  \nonumber \\
\leq& \Lcal_n(W^{(t)}) + b \dotp{\ub \vb^T}{\nabla \Lcal_n(W^{(t)})} + \frac{H b^2}{2} \nonumber \\
\leq& \Lcal_n(W^{(t)}) +  \frac{b\dotp{W^*}{\nabla \Lcal_n(W^{(t)})}}{\norm{W^*}_F} + \frac{H b^2}{2}.
\end{align}
Let $W^{(t)} = UV^T$. Since $V$ is a minimizer of $\Lcal_n(UV^T)$ with
respect to $V$, we have $U^T \nabla \Lcal_n(W^{(t)}) = 0$ and
therefore $ \dotp{W^{(t)}}{\nabla \Lcal_n(W^{(t)})} = {\rm trace}(V U^T \nabla \Lcal_n(W^{(t)}) ) = 0$.
From  convexity of $\Lcal_n(\cdot)$, we have
\begin{align*}
\dotp{W^*}{ \nabla \Lcal_n(W^{(t)}) } 
=& \dotp{W^* - W^{(t)}}{ \nabla \Lcal_n (W^{(t)}) } \\
\leq& \Lcal_n(W^*) - \Lcal_n(W^{(t)}).
\end{align*}
Combining with the display above
\begin{align*}
\Lcal_n(W^{(t)}) - \Lcal_n(W^{(t+1)})  
\geq& \frac{b ( \Lcal_n(W^{(t)}) - \Lcal_n(W^*))}{\norm{W^*}_F} \\
&- \frac{H b^2}{2}.
\end{align*}
By choosing
\[
b = \frac{\Lcal_n(W^{(t)}) - \Lcal_n(W^*) }{H \norm{W^*}_F }
\]
we have
\begin{align*}
\Lcal_n(W^{(t)}) - \Lcal_n(W^{(t+1)}) \geq& \frac{\rbr{\Lcal_n(W^{(t)}) - \Lcal_n(W^*) }^2}{2 H \norm{W^*}_F^2 } \\
\geq& \frac{\rbr{\Lcal_n(W^{(t)}) - \Lcal_n(W^*) }^2}{2 m H A^2 }.
\end{align*}
Using Lemma \ref{lemma:recursion} in Appendix we know that after 
\removed{
\mcomment{It is not clear how you use Lemma B.2. What is $\epsilon_t$, etc?}
}
\[
t \geq \left\lceil \frac{2 m H A^2}{\varepsilon} \right\rceil
\]
iterations, we have $\Lcal_n(W^{(t)}) \leq \Lcal_n(W^*) + \varepsilon$.
\end{proof}

\section{An auxiliary lemma}
\begin{lemma}
(Lemma B.2 of \cite{Shalev-Shwartz2010Trading}) Let $x > 0$ and let $\varepsilon_0,\varepsilon_1,...$ be a sequence such that $\varepsilon \leq \varepsilon_t - r \varepsilon_t^2$ for all $t$. Let $\varepsilon$ be a positive scalar and $t$ be a positive integer such that $t \geq \lceil \frac{1}{x\varepsilon} \rceil$. Then $\varepsilon_t \leq \varepsilon$.
\label{lemma:recursion}
\end{lemma}

\section{Evaluation on Real World Datasets}
\label{sec:realworld}

\begin{figure*}[t]
\begin{center}
\includegraphics[width=0.33 \textwidth]{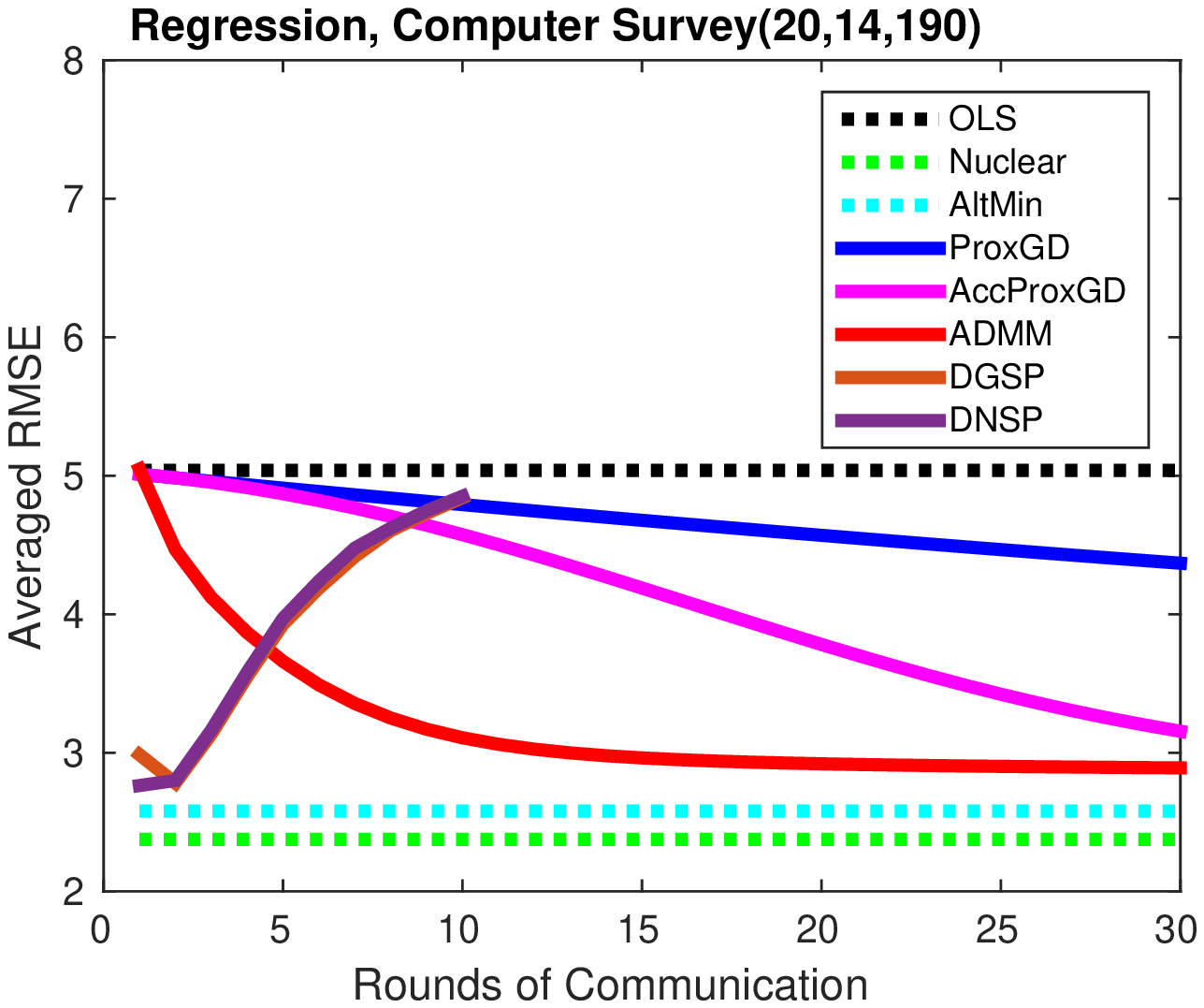}%
\includegraphics[width=0.33 \textwidth]{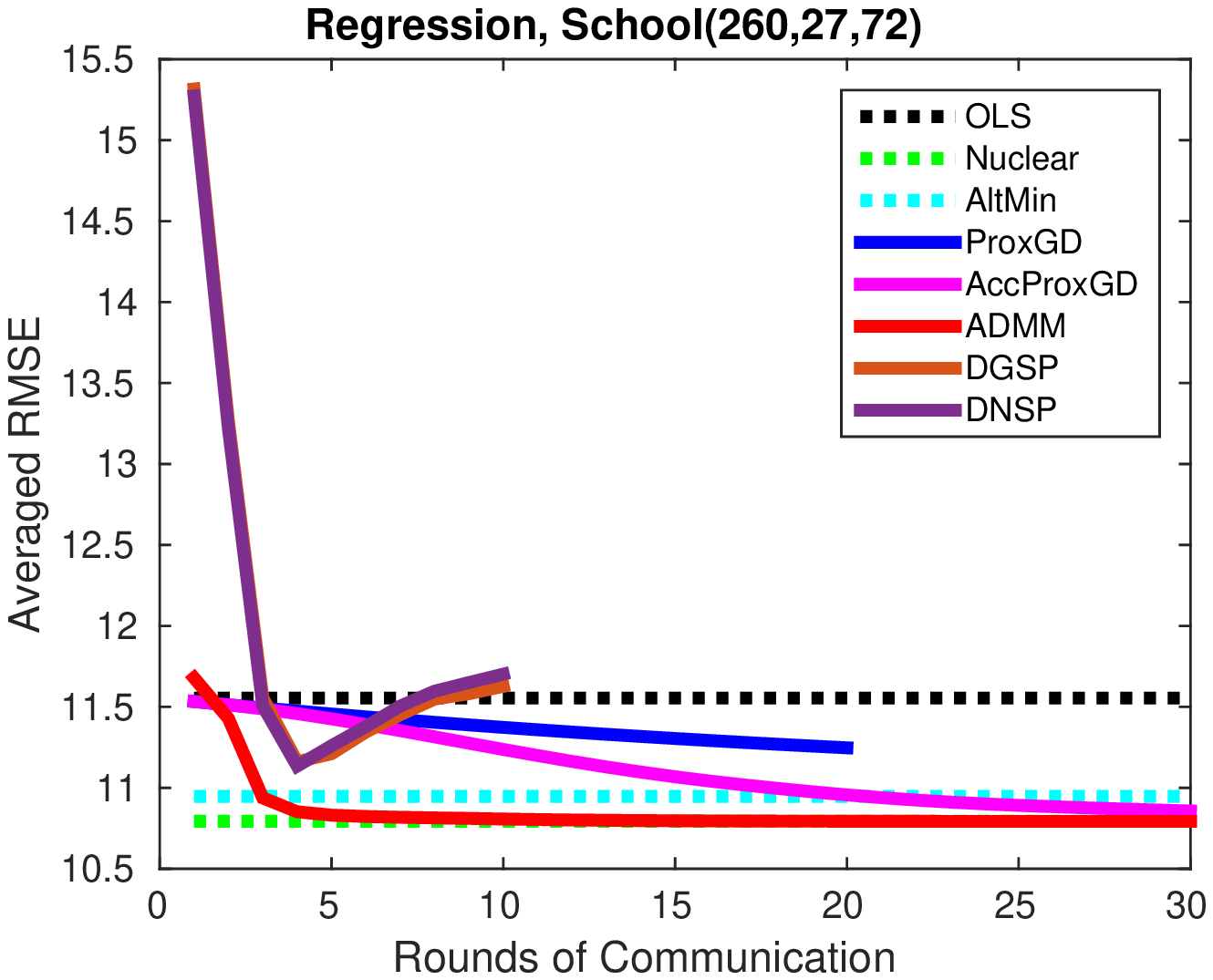}%
\includegraphics[width=0.33 \textwidth]{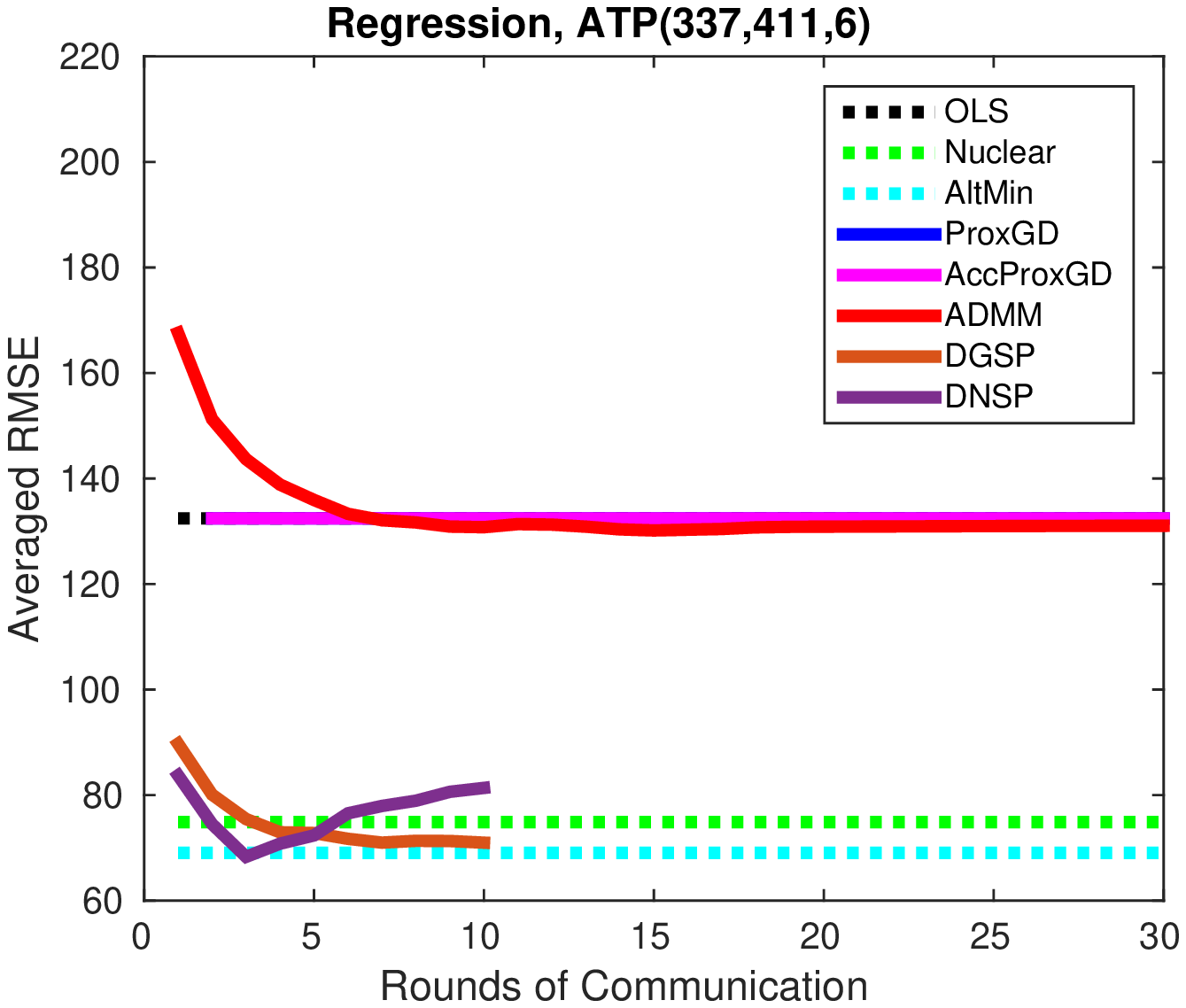}%
\end{center}
\begin{center}
\includegraphics[width=0.33 \textwidth]{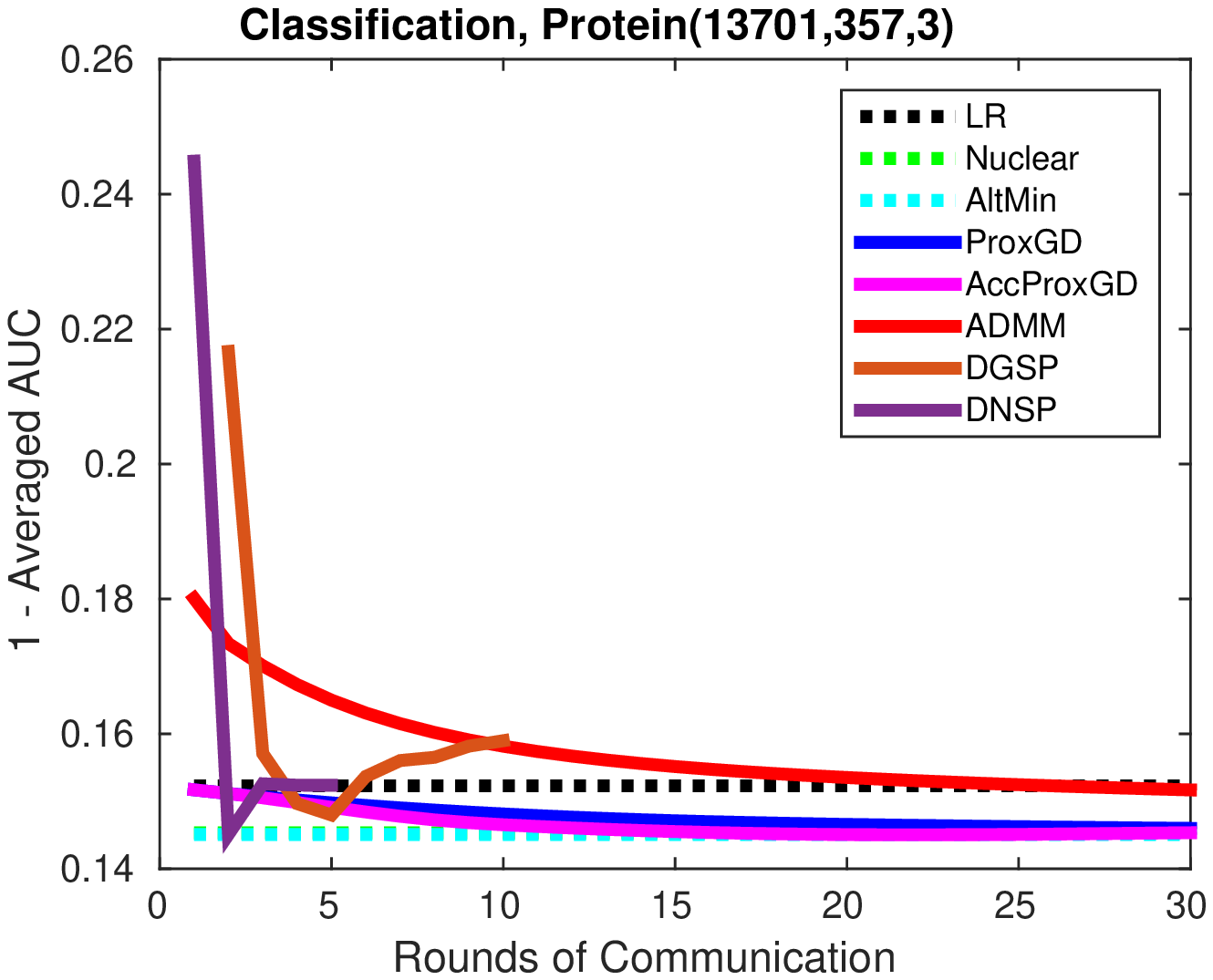}%
\includegraphics[width=0.33 \textwidth]{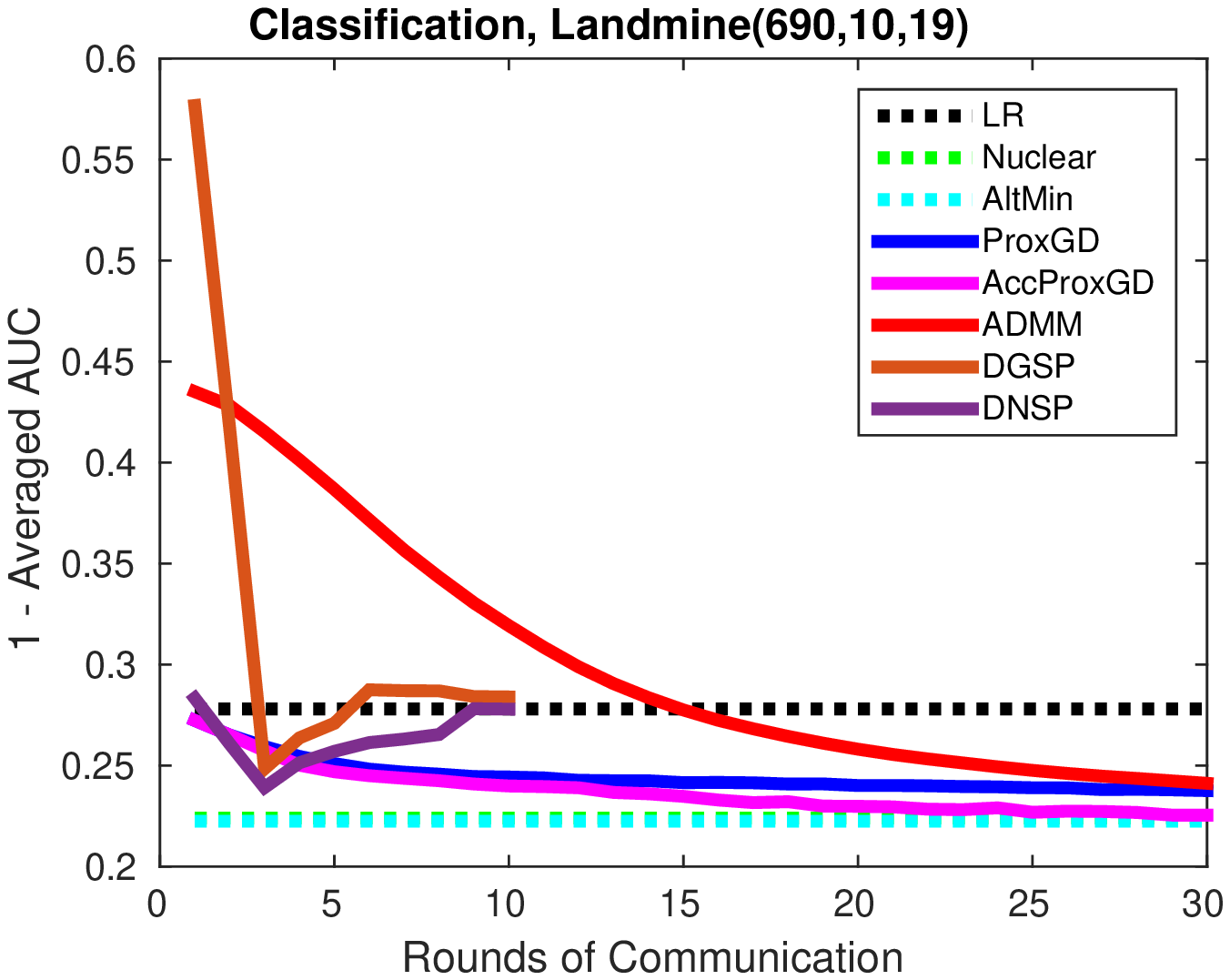}%
\includegraphics[width=0.33 \textwidth]{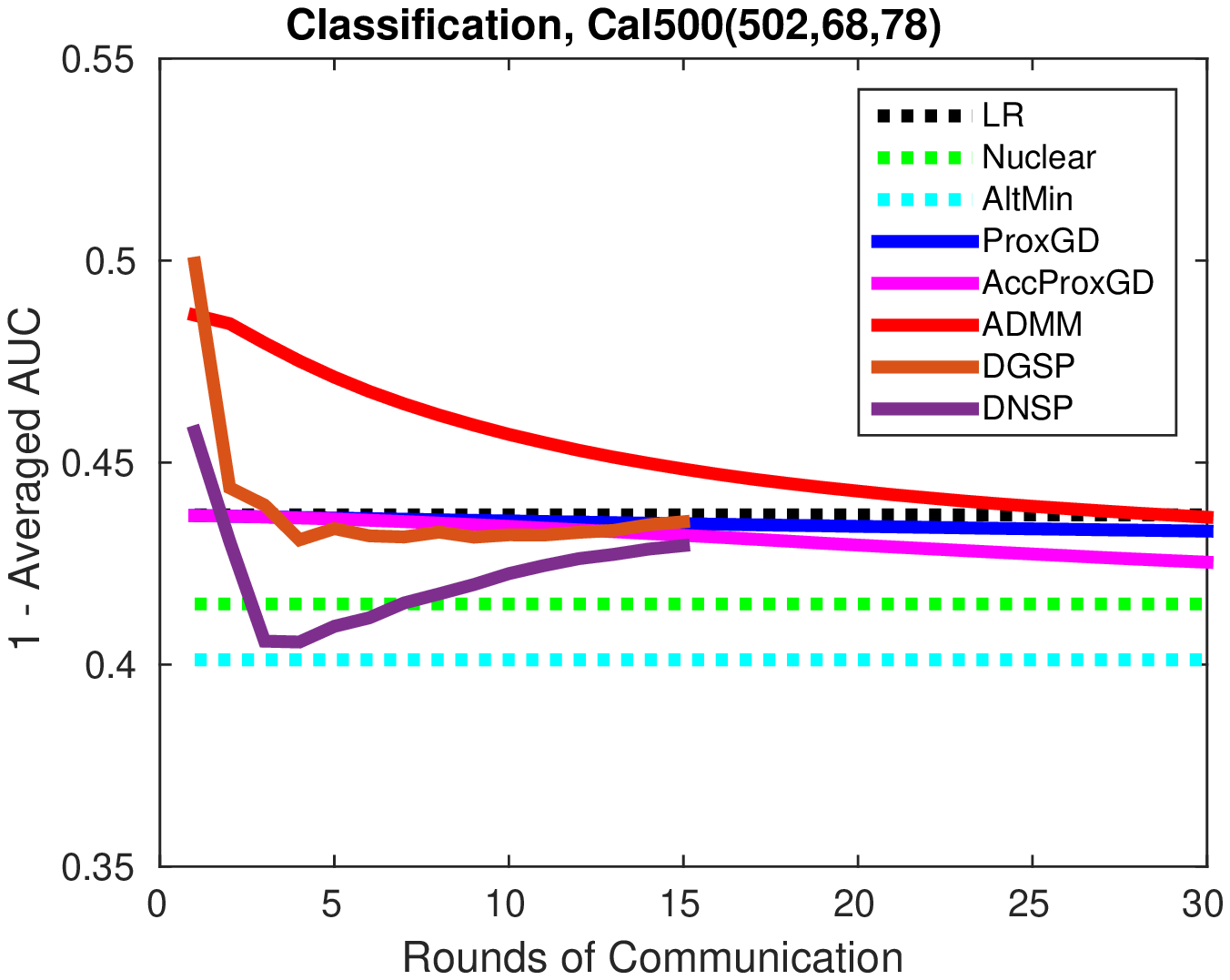}%
\end{center}
\caption{Prediction Error on real data.}
\label{fig:real_data}
\end{figure*}

We also evaluate discussed algorithms on several real world data sets,
with $20\%$ of the whole dataset as training set, $20\%$ as held-out
validation, then report the testing performance on the remaining
$60\%$. For the real data, we have observed that adding $\ell_2$
regularization usually helps improving the generalization
performance. For the \texttt{Local} procedure we added an $\ell_2$
regularization term \removed{\mcomment{maybe write an objective to be
  minimized}} (leads to ridge regression or $\ell_2$ regularized
logistic regression).  For \texttt{DGSP} and \texttt{DNSP}, we also
add an $\ell_2$ regularization in finding the subspaces and refitting
\removed{\mcomment{Again add details}}.  We have worked on the following
multi-task learning datasets\removed{\mcomment{You write ``datasets'', ``data-sets'' and ``data sets'' throughout that paper. This should be consistent.}}:

\textbf{School.\footnote{\url{http://cvn.ecp.fr/personnel/andreas/code/mtl/index.html}}}
The dataset consists of examination scores of students from London's
secondary schools during the years 1985, 1986, 1987. There are 27
school-specific and student-specific features to describe each
student. The instances are divided by different schools, and the task is to
predict the students' performance. We only considered schools with at
least $100$ records, which results in 72 tasks in total. The maximum
number of records for each individual school is 260.

\textbf{Computer Survey.} The data is taken from a conjoint analysis
experiment \citep{lenk1996hierarchical} which surveyed 180 persons
about the probability of purchasing 20 kinds of personal
computers. There are 14 variables for each computer, the response is
an integer rating with scale $0-10$.

\textbf{ATP.\footnote{\url{http://mulan.sourceforge.net/datasets.html}}}
The task here is to predict the airline ticket price
\citep{Spyromitros-Xioufis2012Multi}. We are interested in the minimum
prices next day for some specific observation date and departure date
pairs. Each case is described by 411 features, and there are 6 target
minimum prices for different airlines to predict. The sample size is
337.

\textbf{Protein.} Given the amino acid sequence, we are interested
predicting the protein secondary structure
\citep{Sander1991Database}. We tackle the problem by considering the
following three binary classification tasks: coil vs helix, helix vs
strand, strand vs coil. Each sequence is described by 357
features. There are 24,387 instances in total.

\textbf{Landmine.} The data is collected from 19 landmine detection
tasks \citep{xue2007multi}. Each landmine field is represented by a
9-dimensional vector extracted from radar images, containing
moment-based, correlation-based, energy ratio, and spatial variance
features. The sample size for each task varies from 445 to 690.

\textbf{Cal500.\footnote{\url{http://eceweb.ucsd.edu/~gert/calab/}}}
This music dataset \citep{turnbull2008semantic} consists of 502 songs,
where for each song 68 features are extracted. Each task is to predict
whether a particular musically relevant semantic keyword should be
an annotation for the song. We only consider tags with at least 50 times
apperance, which results in 78 prediction tasks.

We compared various approaches as in the simulation study, except the
\texttt{BestRep} as the best low-dimensional representation is
unknown. We also compared with \texttt{AltMin}, which learns low-rank
prediction matrix using the alternating minimization
\citep{jain2013low}.  The results are shown in Figure
\ref{fig:real_data}. Since the labels for the real world
classification datasets are often unbalanced, we report averaged area
under the curve (AUC) instead of classification accuracy. We have the
following observations:
\begin{itemize}
\item The distributed first-order approaches converge much slower
  than in simulations, especially on ATP and Cal500. We
  suspect this is because in the simulation study, the generated data 
  are usually well conditioned, which makes faster convergence possible
  for such methods \citep{Agarwal2012Fast,hong2012linear}. On
  real data, the condition number can be much worse.

\item In most case, \texttt{DNSP} is the best in terms of
  communication-efficiency. \texttt{DGSP} also has reasonable
  performance with fewer round of communications compared to
  distributed first-order approaches.

\item Among the first-order distributed convex optimization methods,
  \texttt{AccProxGD} is overall the most communication-efficient,
  while \texttt{DFW} is the worst, though it might have some
  advantages in terms of computation. Also, we observed significant
  zig-zag behavior of the \texttt{DFW} algorithm, as discussed in
  \citep{lacoste2015global}.
\end{itemize}

\section{Full experimental results with Distributed Frank-Wolfe}

\begin{figure*}[t]
\begin{center}
\includegraphics[width=0.33 \textwidth]{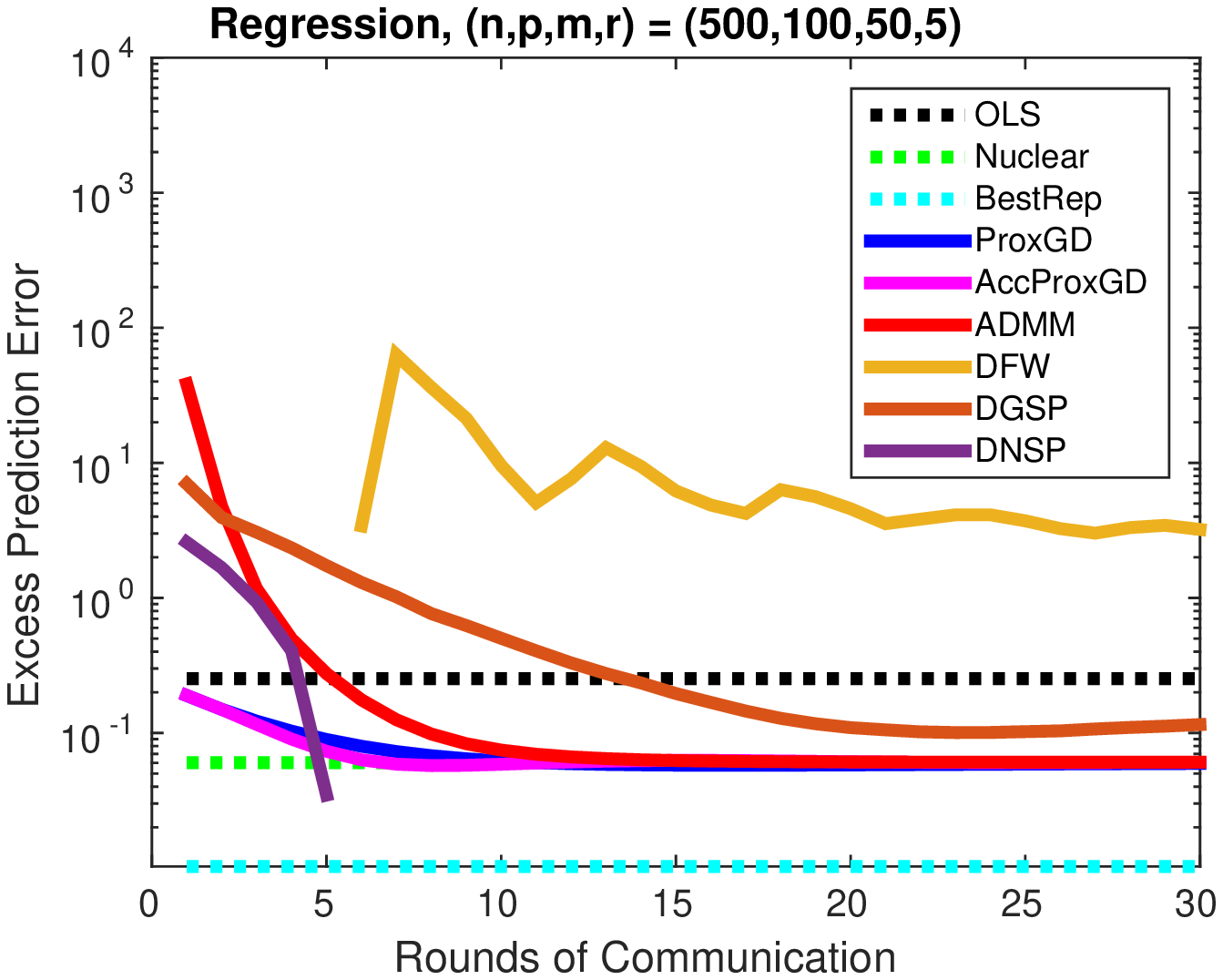}%
\includegraphics[width=0.33 \textwidth]{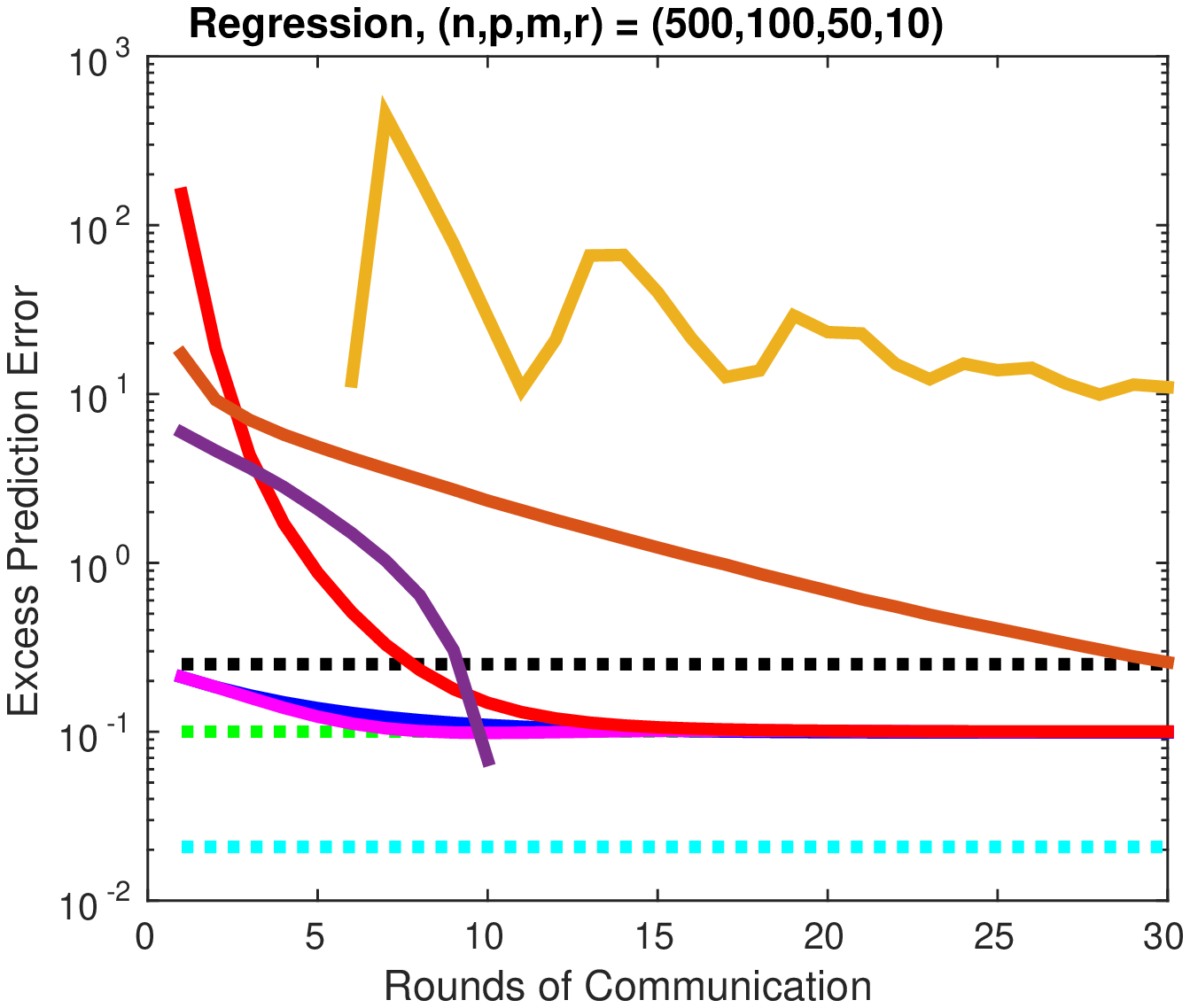}%
\includegraphics[width=0.33 \textwidth]{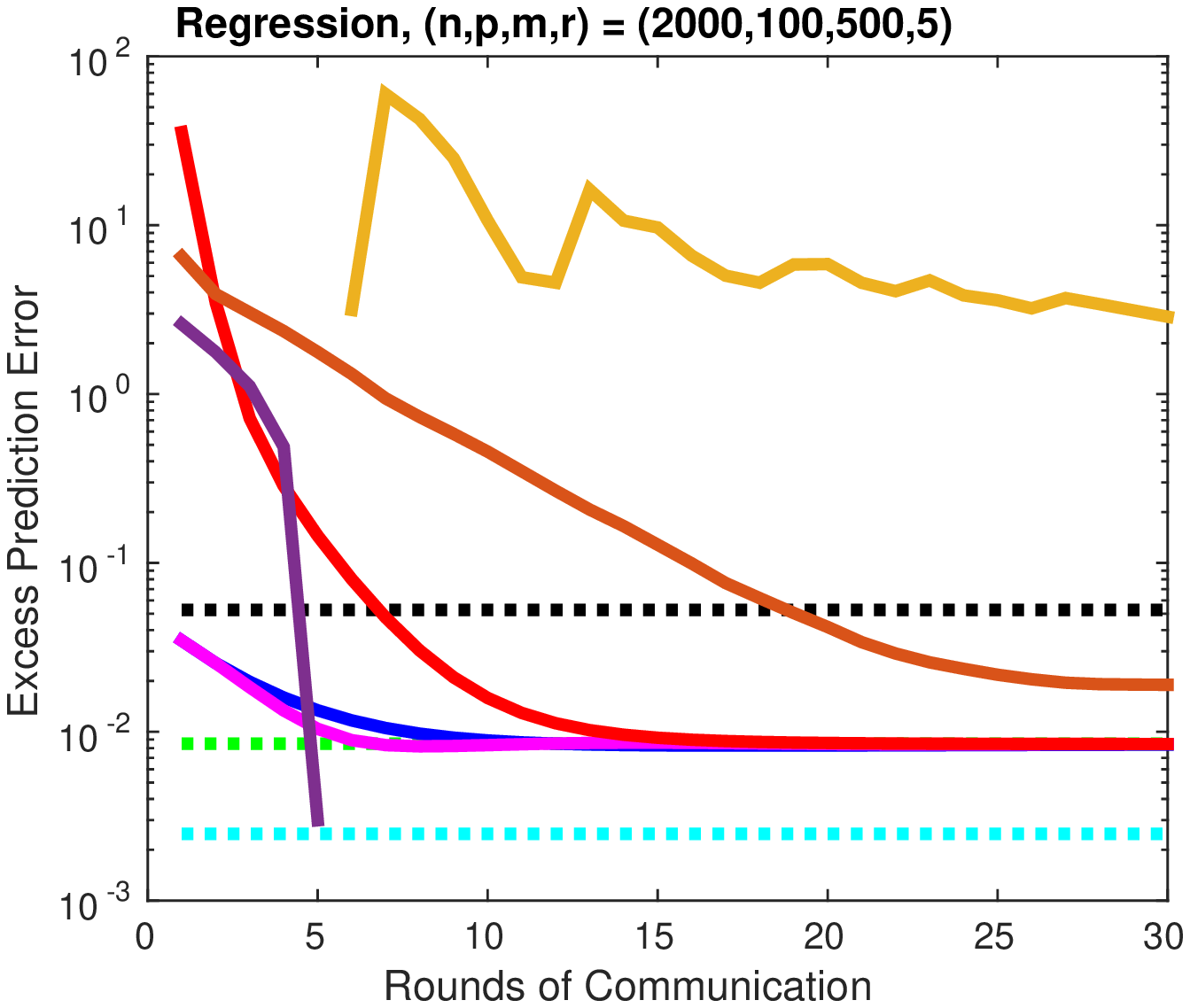}%
\end{center}
\begin{center}
\includegraphics[width=0.33 \textwidth]{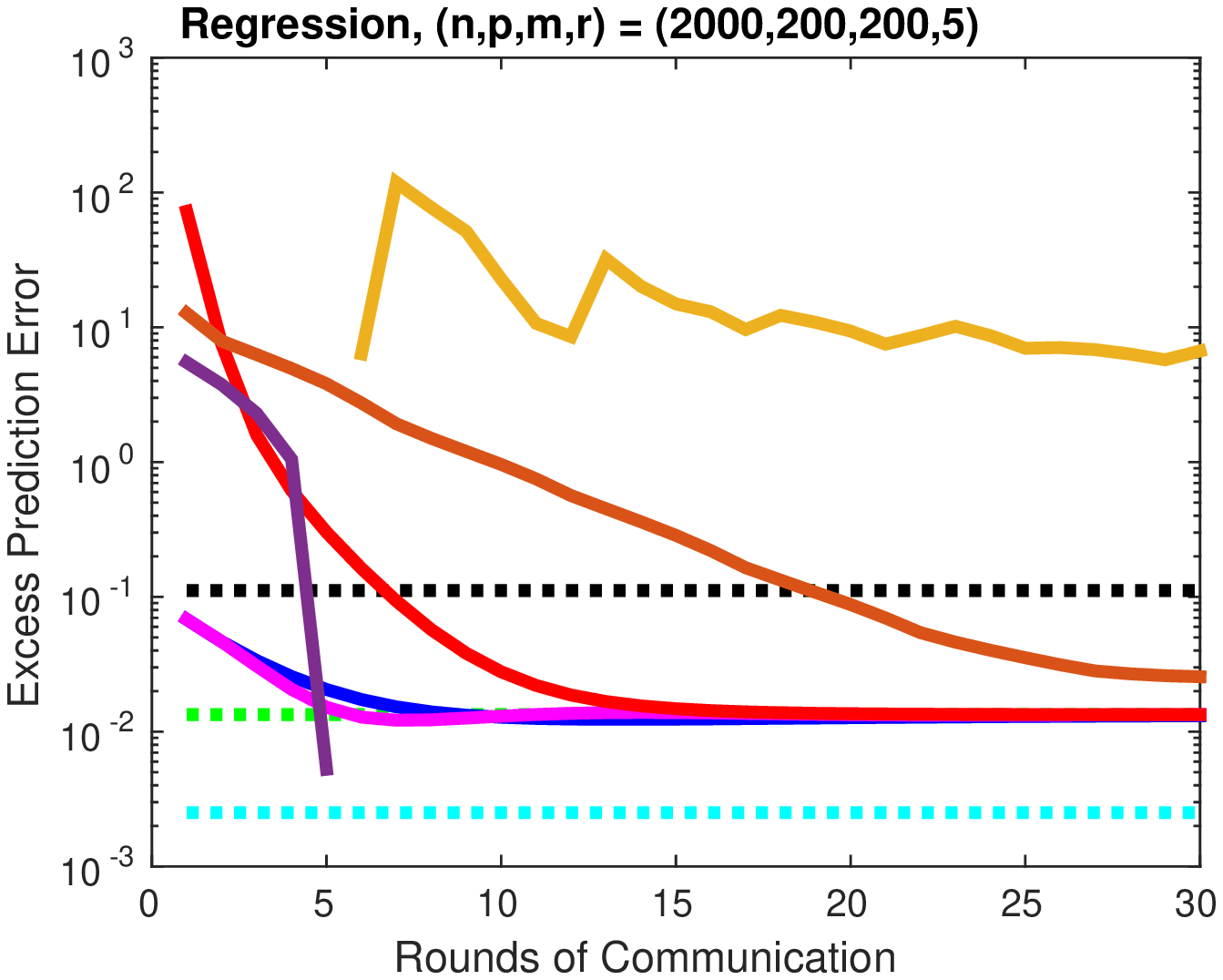}%
\includegraphics[width=0.33 \textwidth]{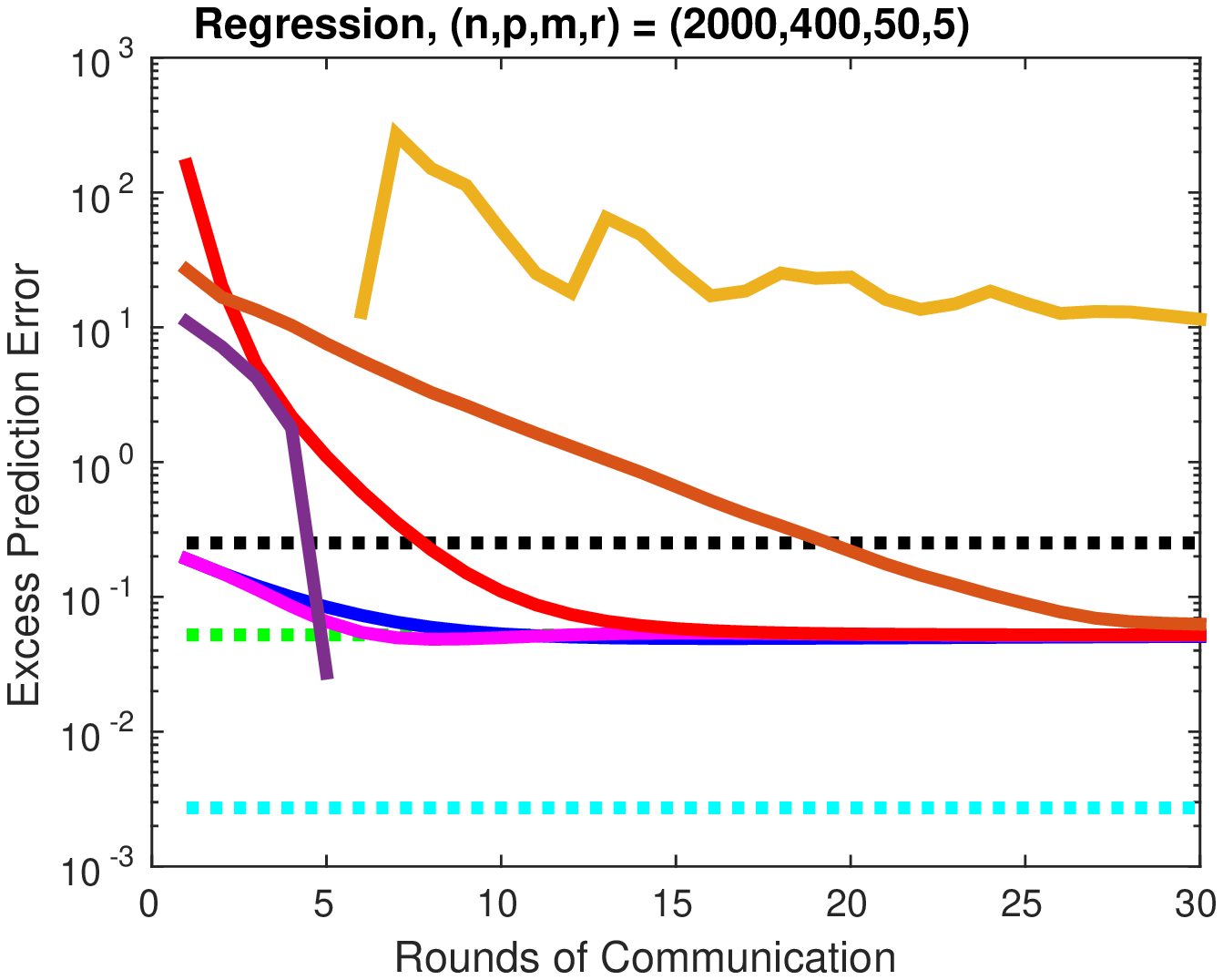}%
\includegraphics[width=0.33 \textwidth]{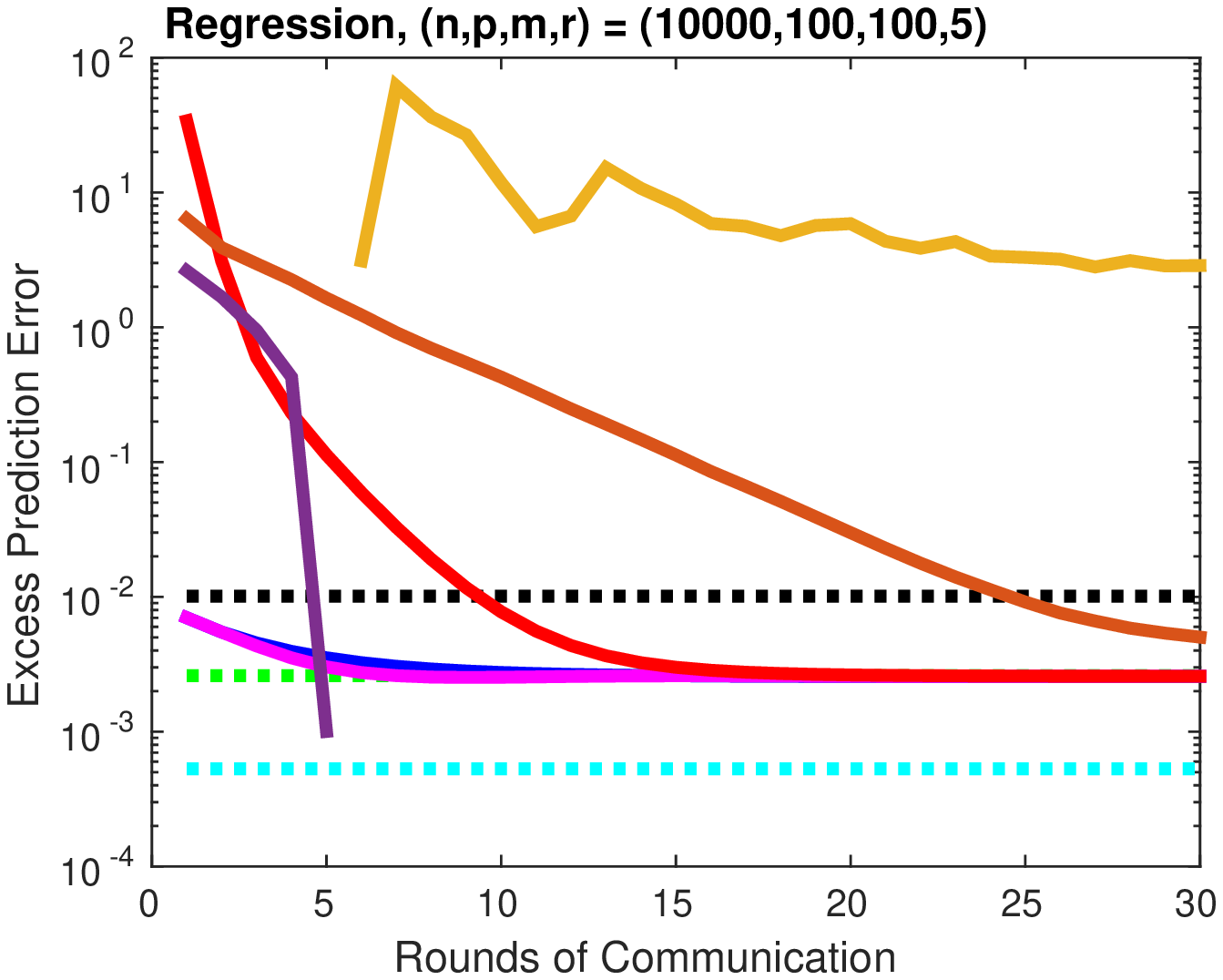}%
\end{center}
\caption{Excess prediction error for
  multi-task regression.}
\label{fig:fw_simulation_regression}
\end{figure*}

\begin{figure*}[t]
\begin{center}
\includegraphics[width=0.33 \textwidth]{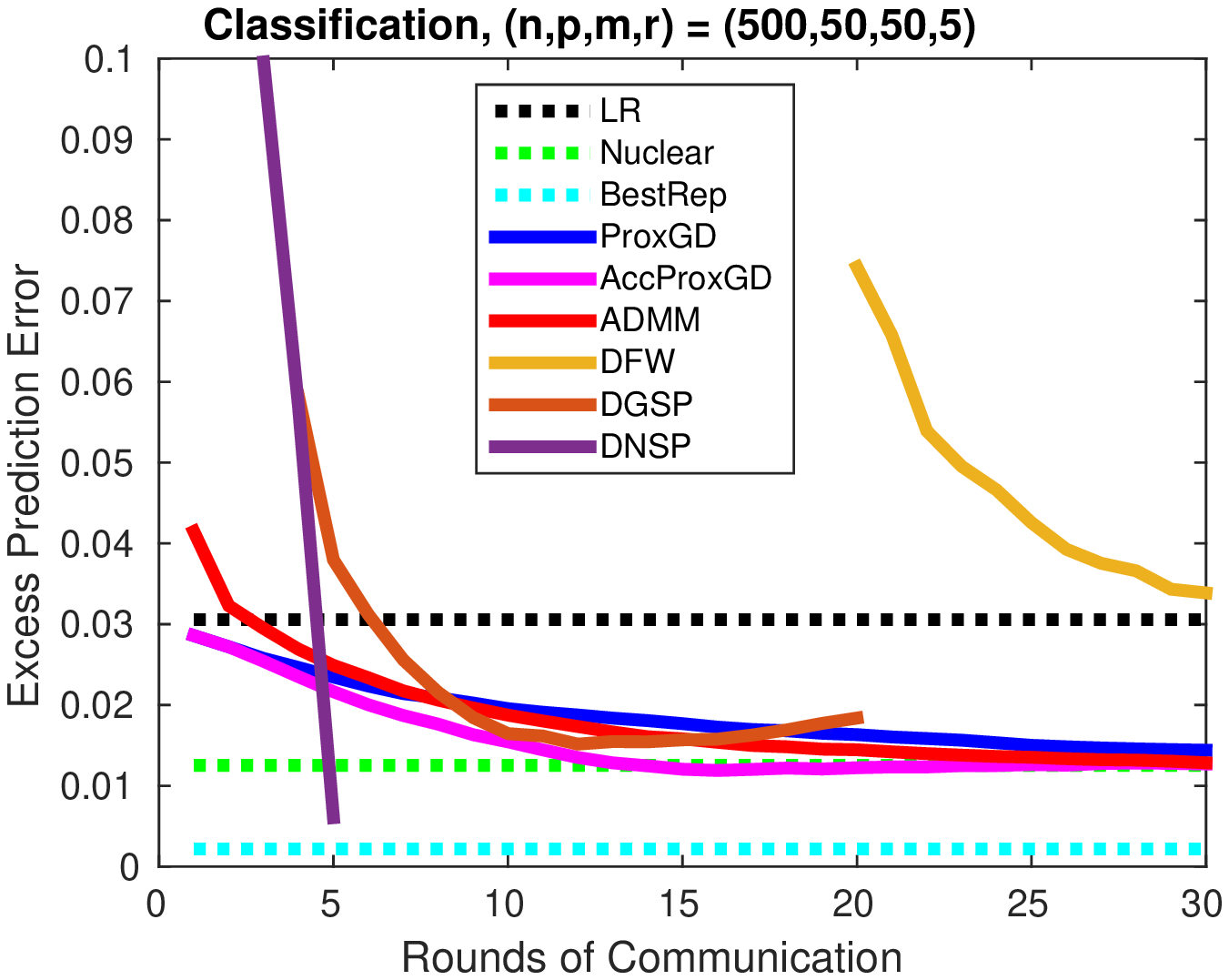}%
\includegraphics[width=0.33 \textwidth]{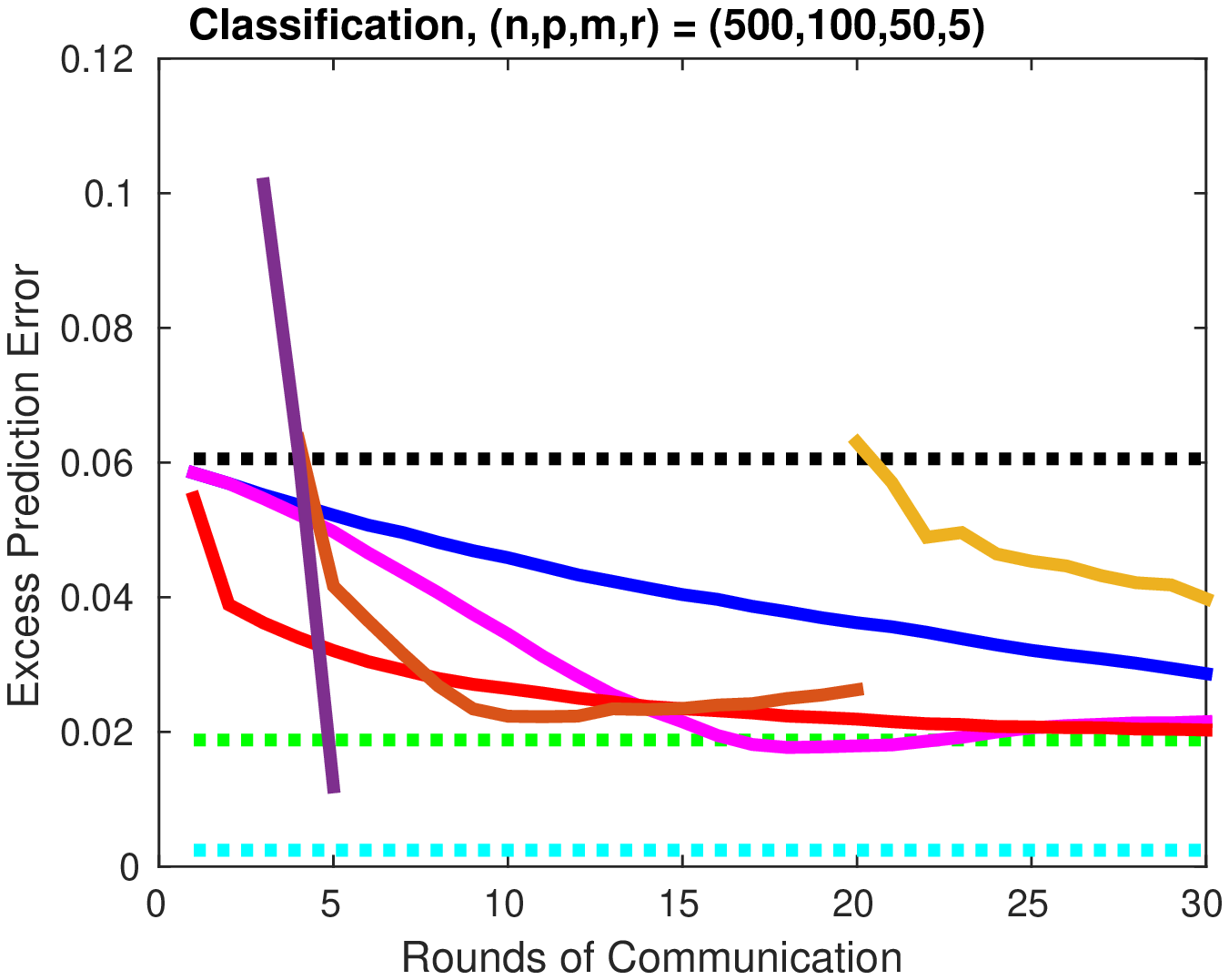}%
\includegraphics[width=0.33 \textwidth]{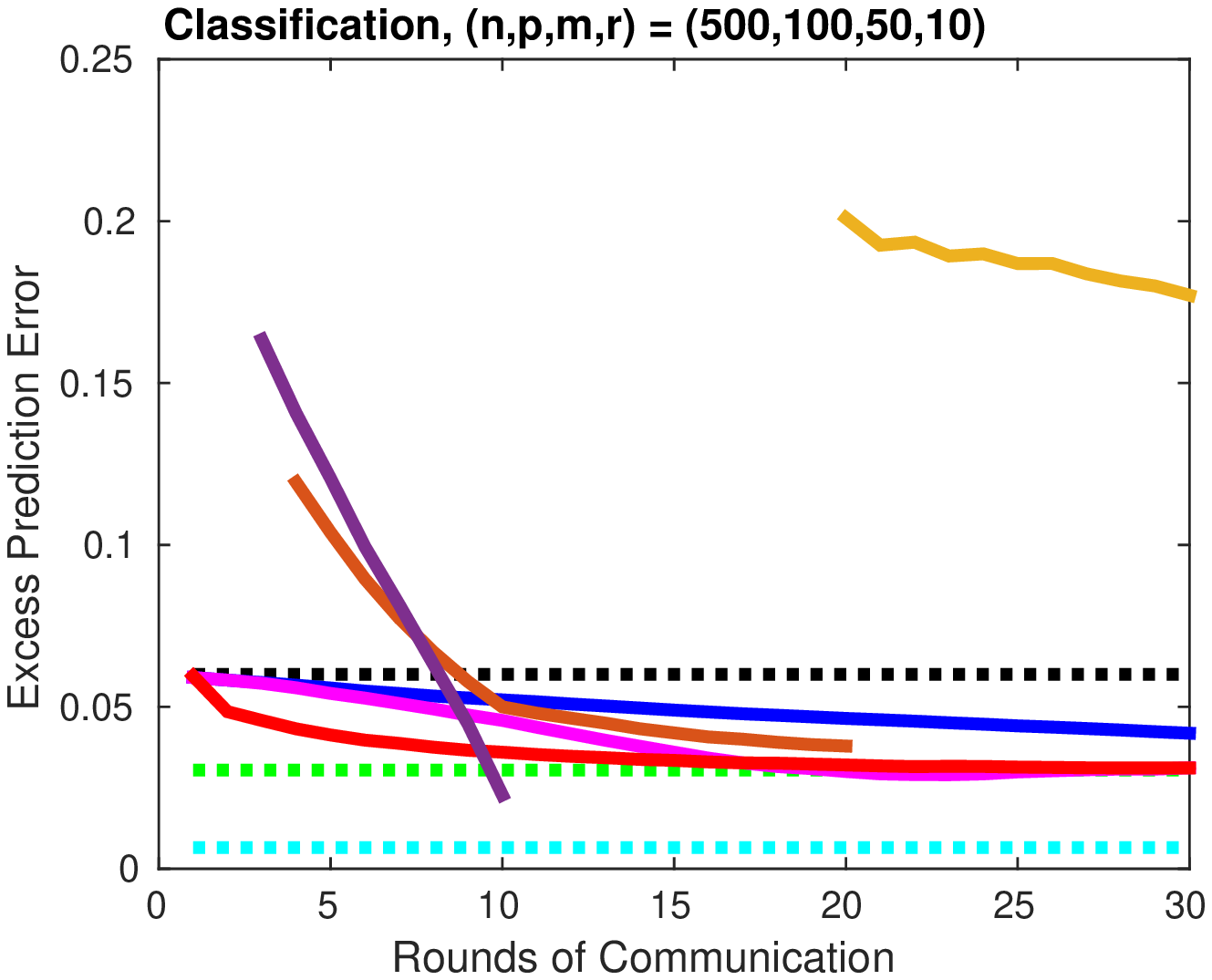}%

\end{center}
\begin{center}
\includegraphics[width=0.33 \textwidth]{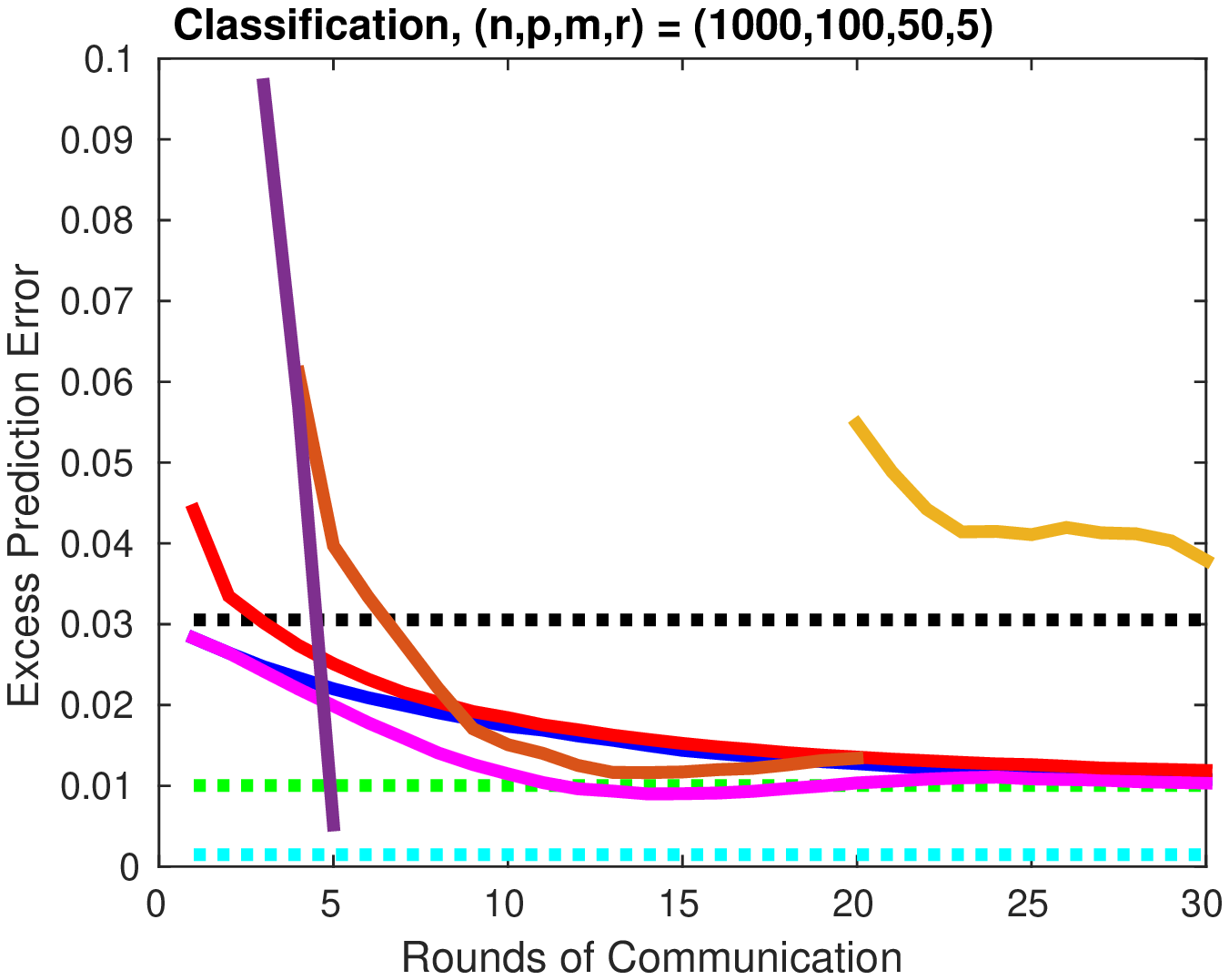}%
\includegraphics[width=0.33 \textwidth]{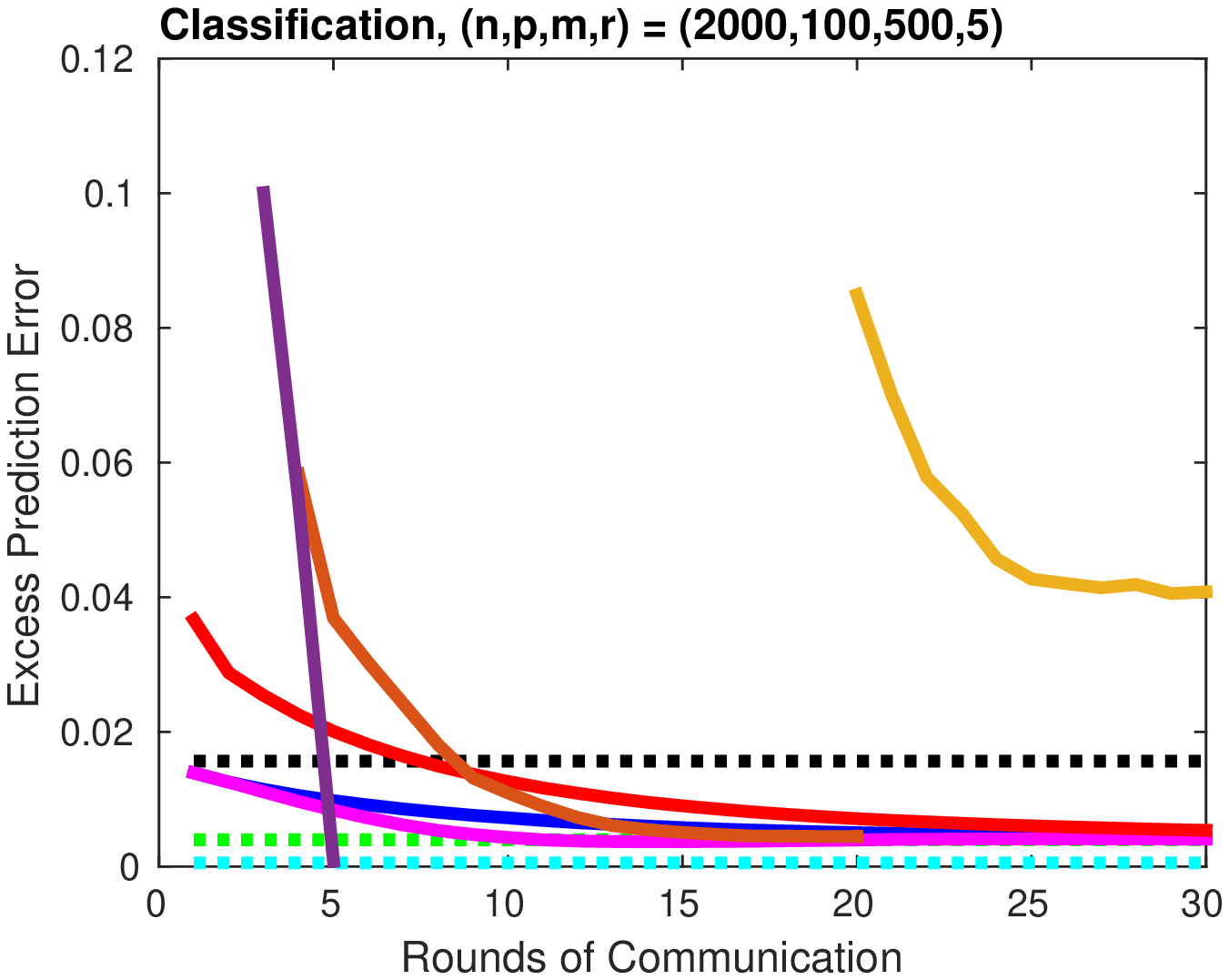}%
\includegraphics[width=0.33 \textwidth]{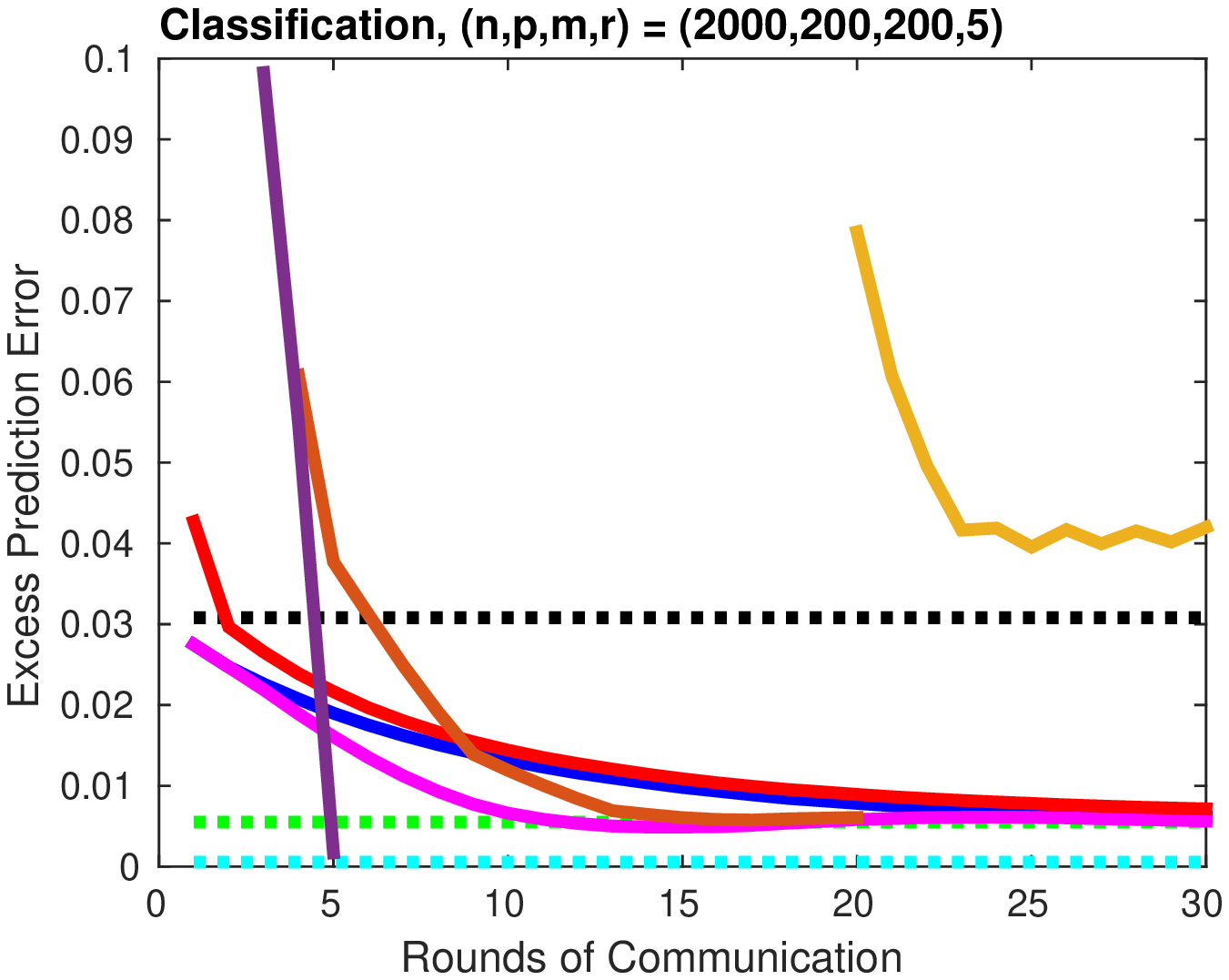}%
\end{center}
\caption{Excess prediction error for
  multi-task classification.}
\label{fig:fw_simulation_classification}
\end{figure*}

\begin{figure*}[t]
\begin{center}
\includegraphics[width=0.33 \textwidth]{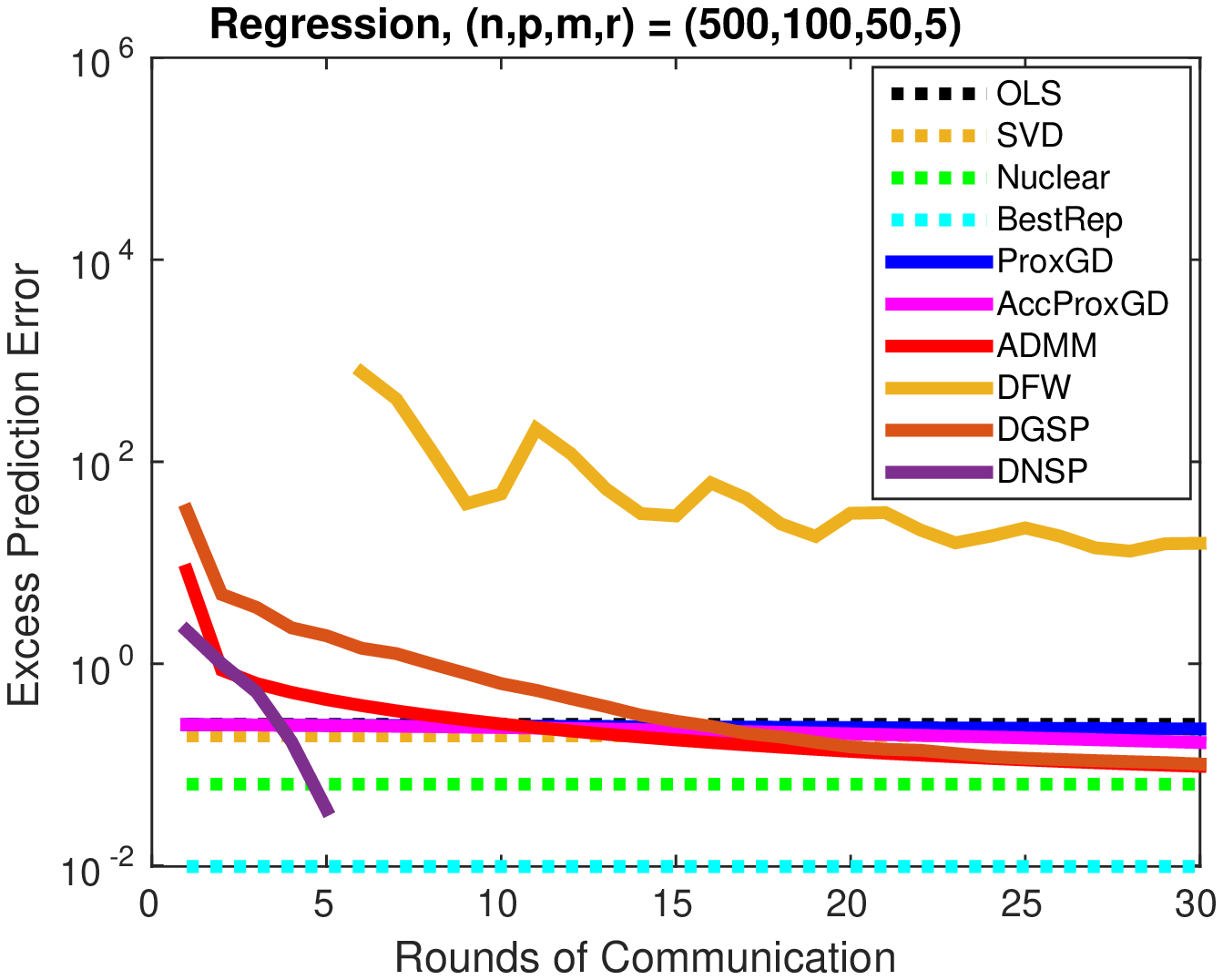}%
\includegraphics[width=0.33 \textwidth]{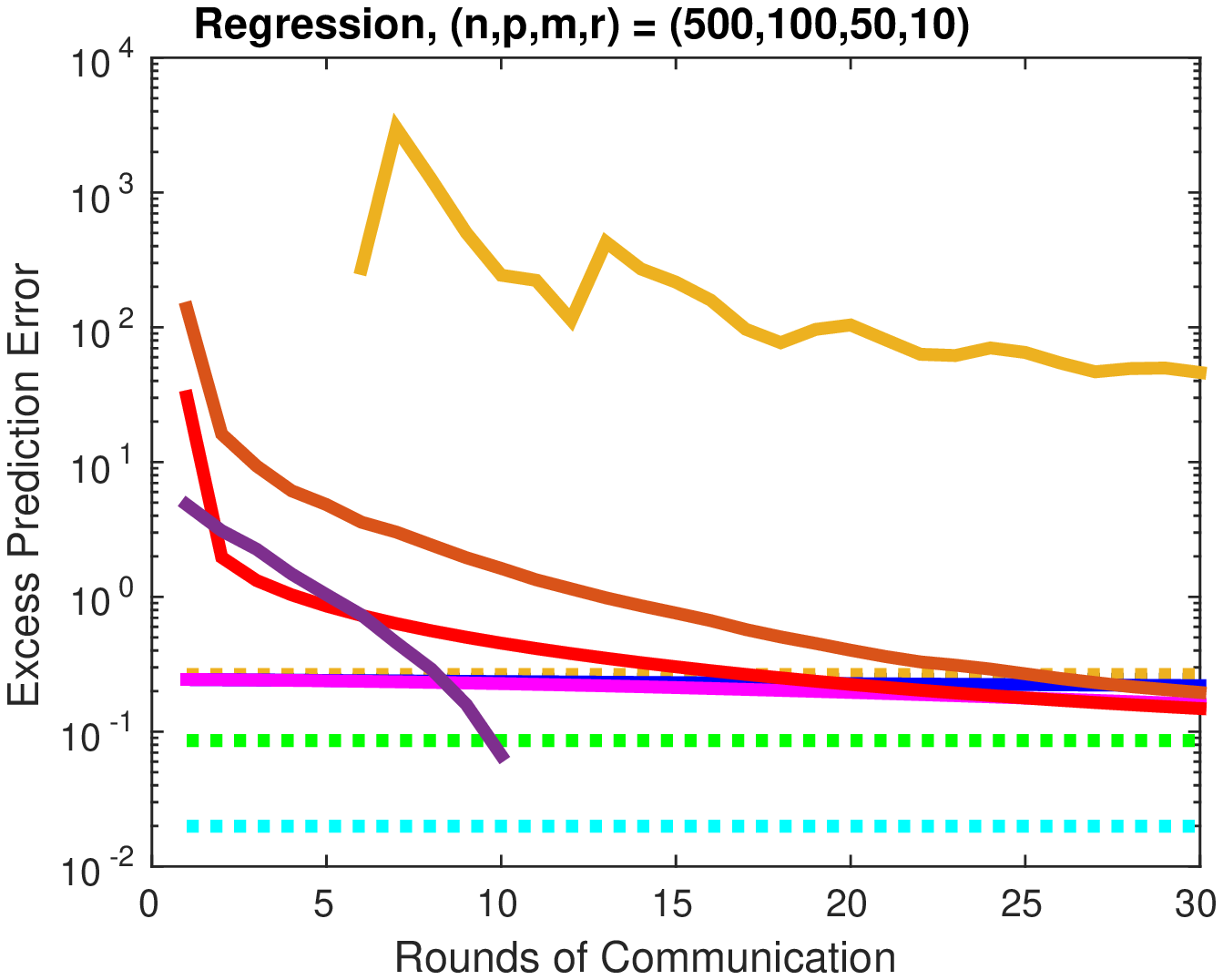}%
\includegraphics[width=0.33 \textwidth]{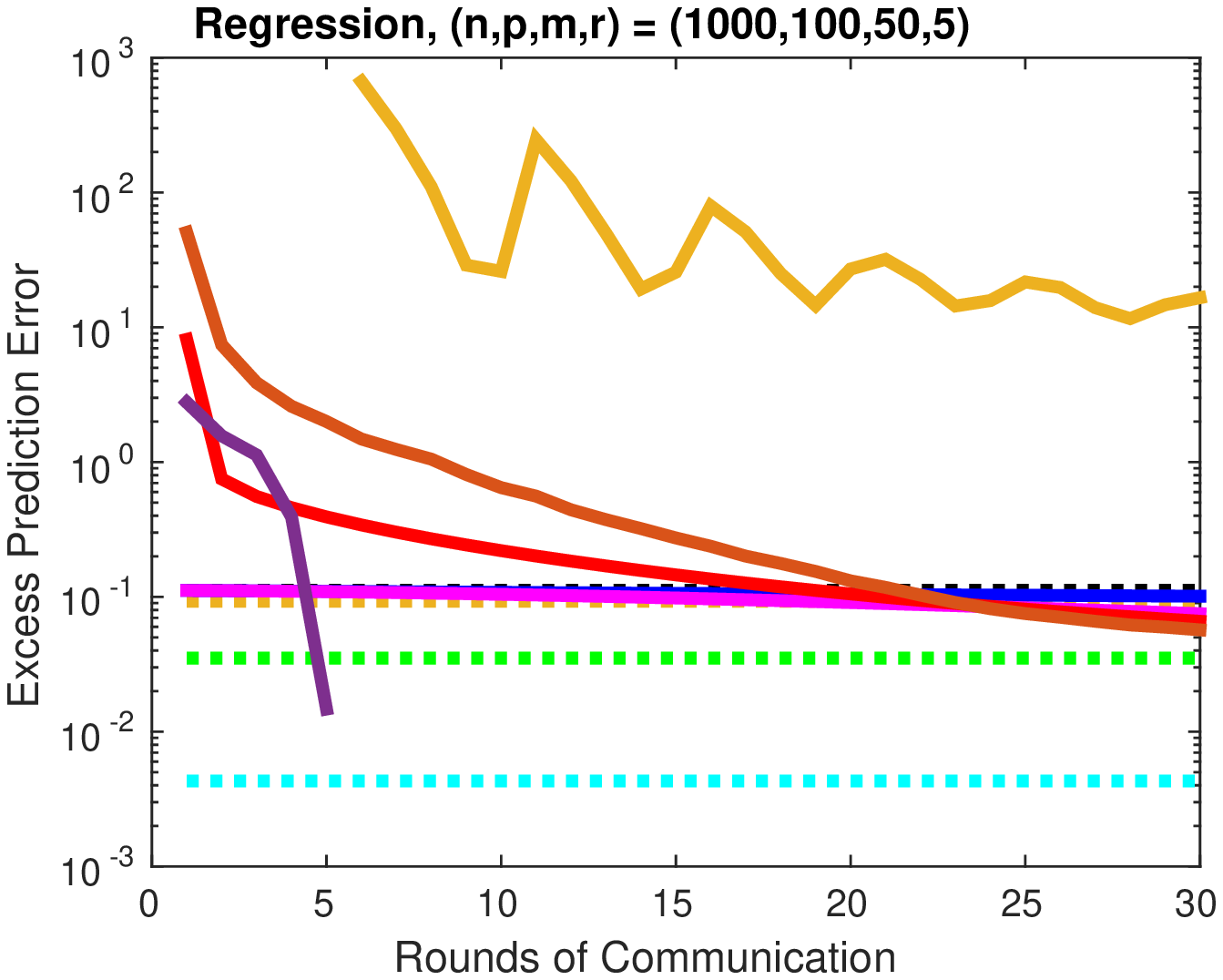}%
\end{center}
\caption{Excess prediction error for
  multi-task regression, with highly correlated features.}
\label{fig:fw_simulation_regression_hard}
\end{figure*}

\begin{figure*}[t]
\begin{center}
\includegraphics[width=0.33 \textwidth]{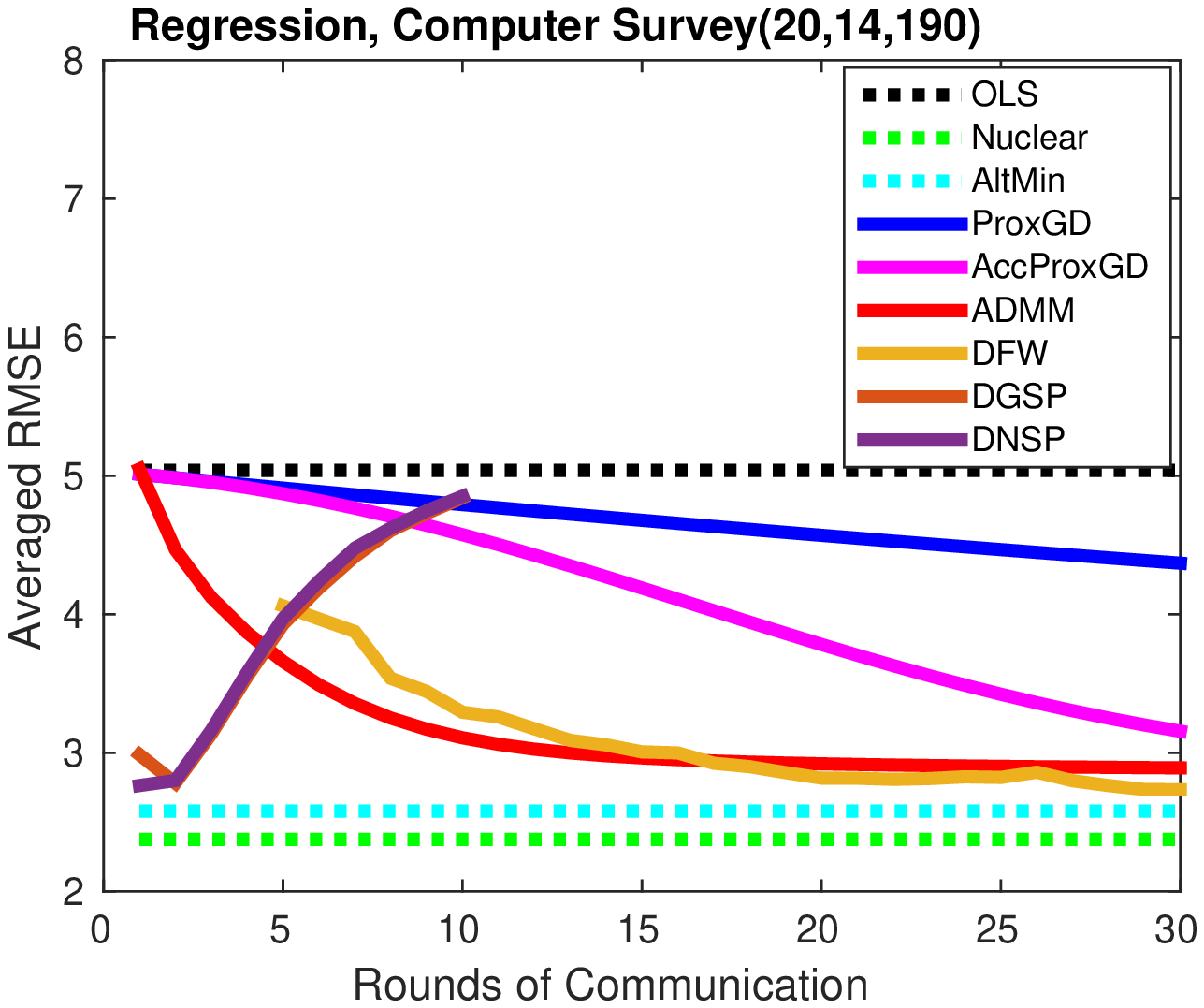}%
\includegraphics[width=0.33 \textwidth]{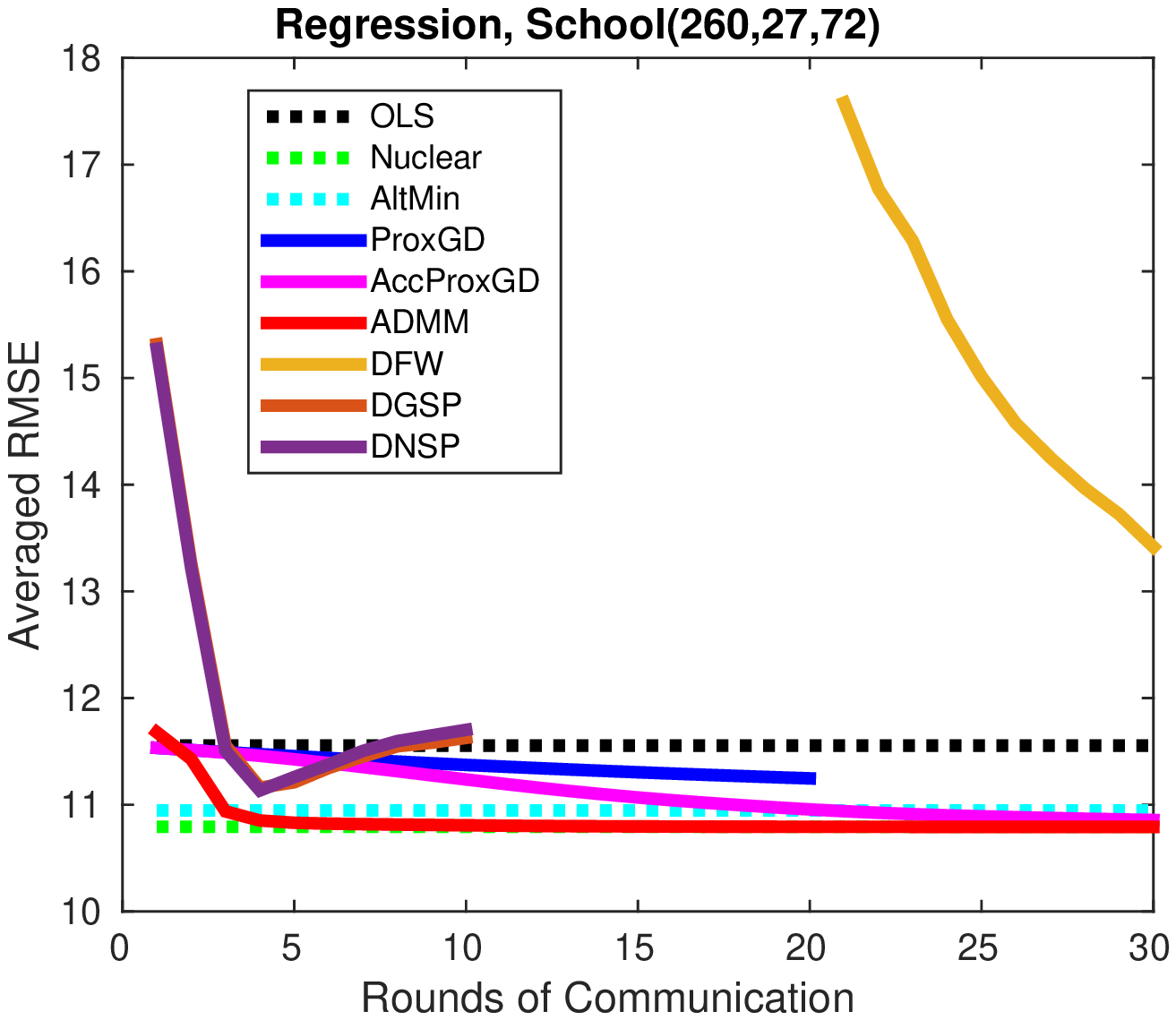}%
\includegraphics[width=0.33 \textwidth]{newfigures/atp.eps}%
\end{center}
\begin{center}
\includegraphics[width=0.33 \textwidth]{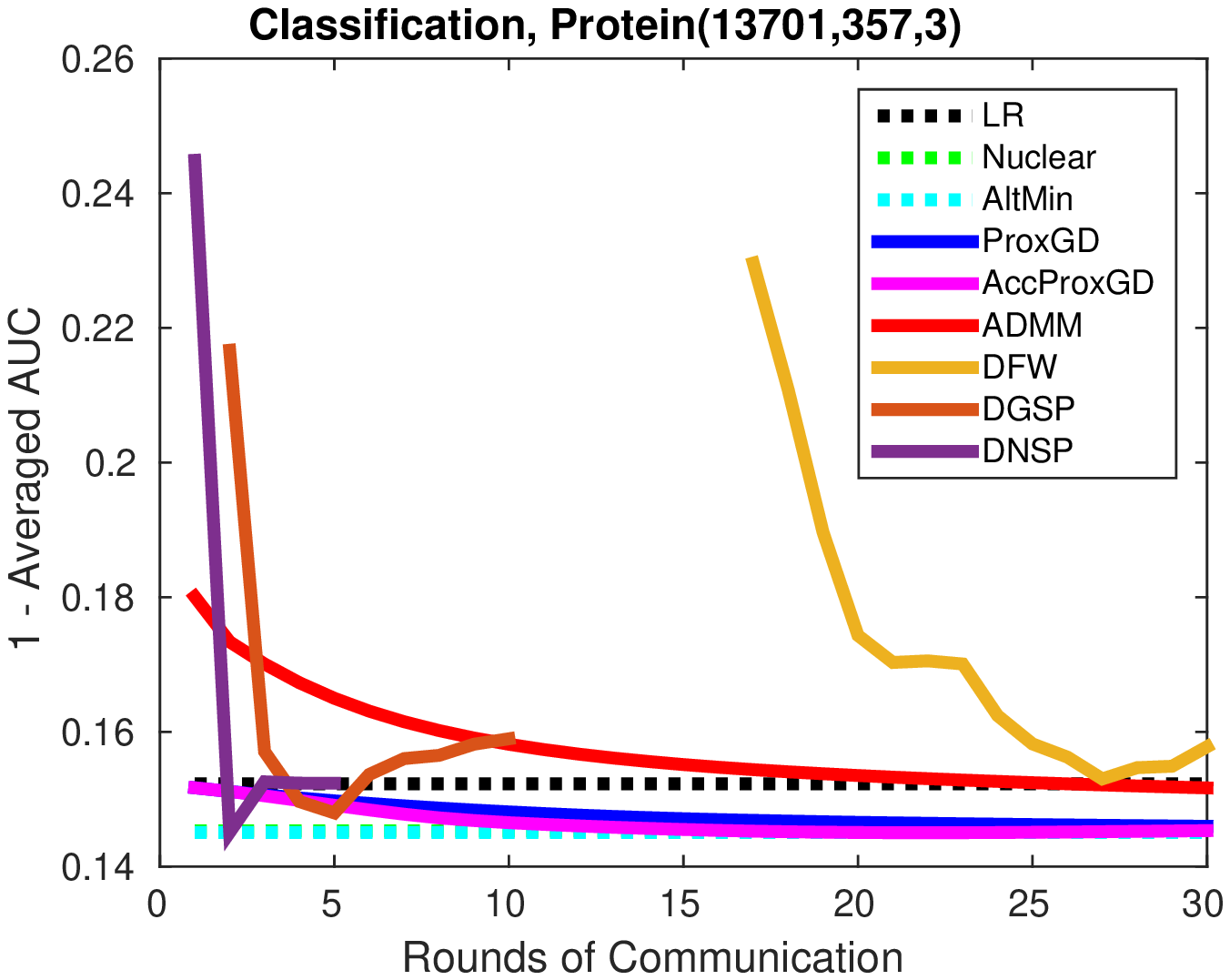}%
\includegraphics[width=0.33 \textwidth]{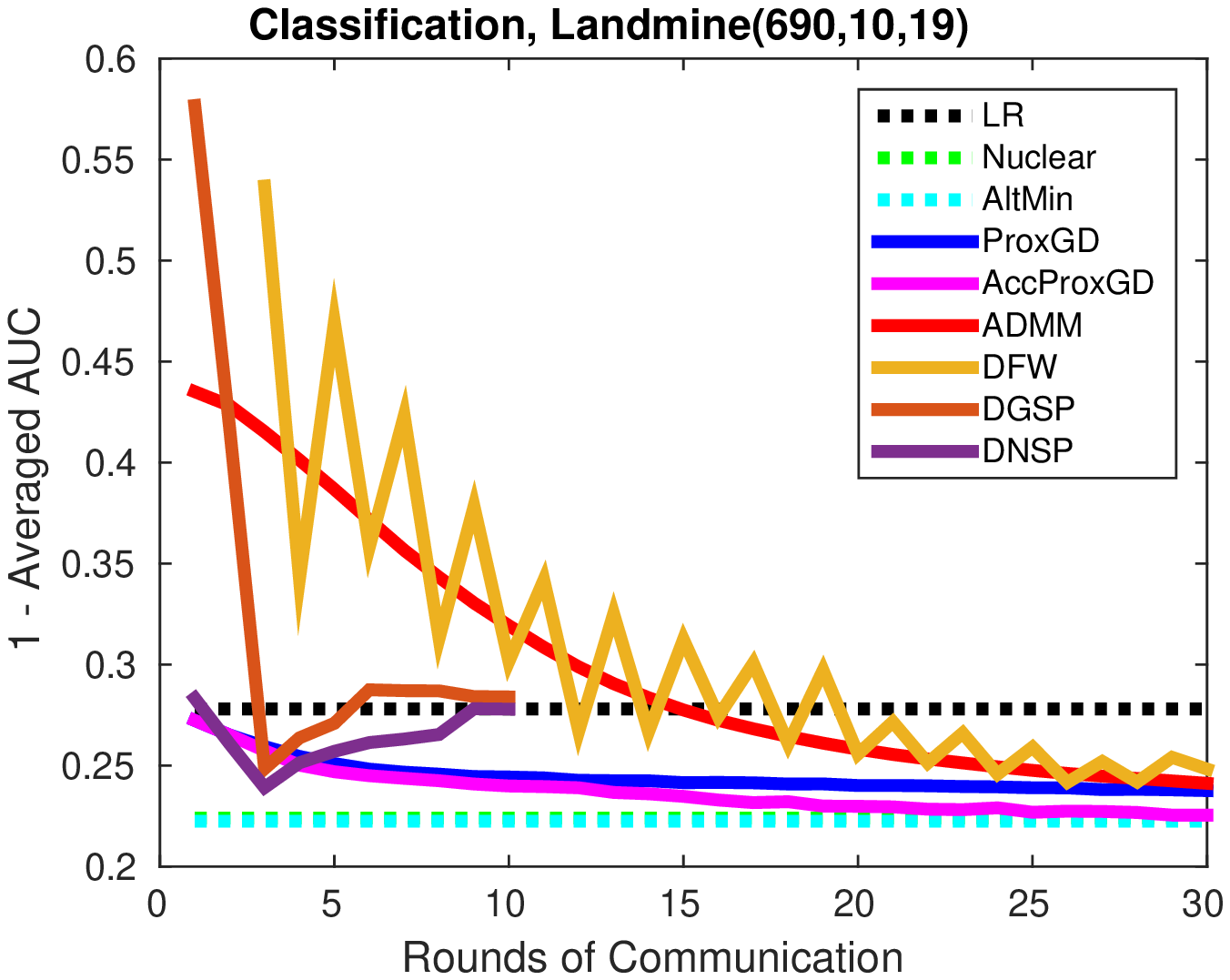}%
\includegraphics[width=0.33 \textwidth]{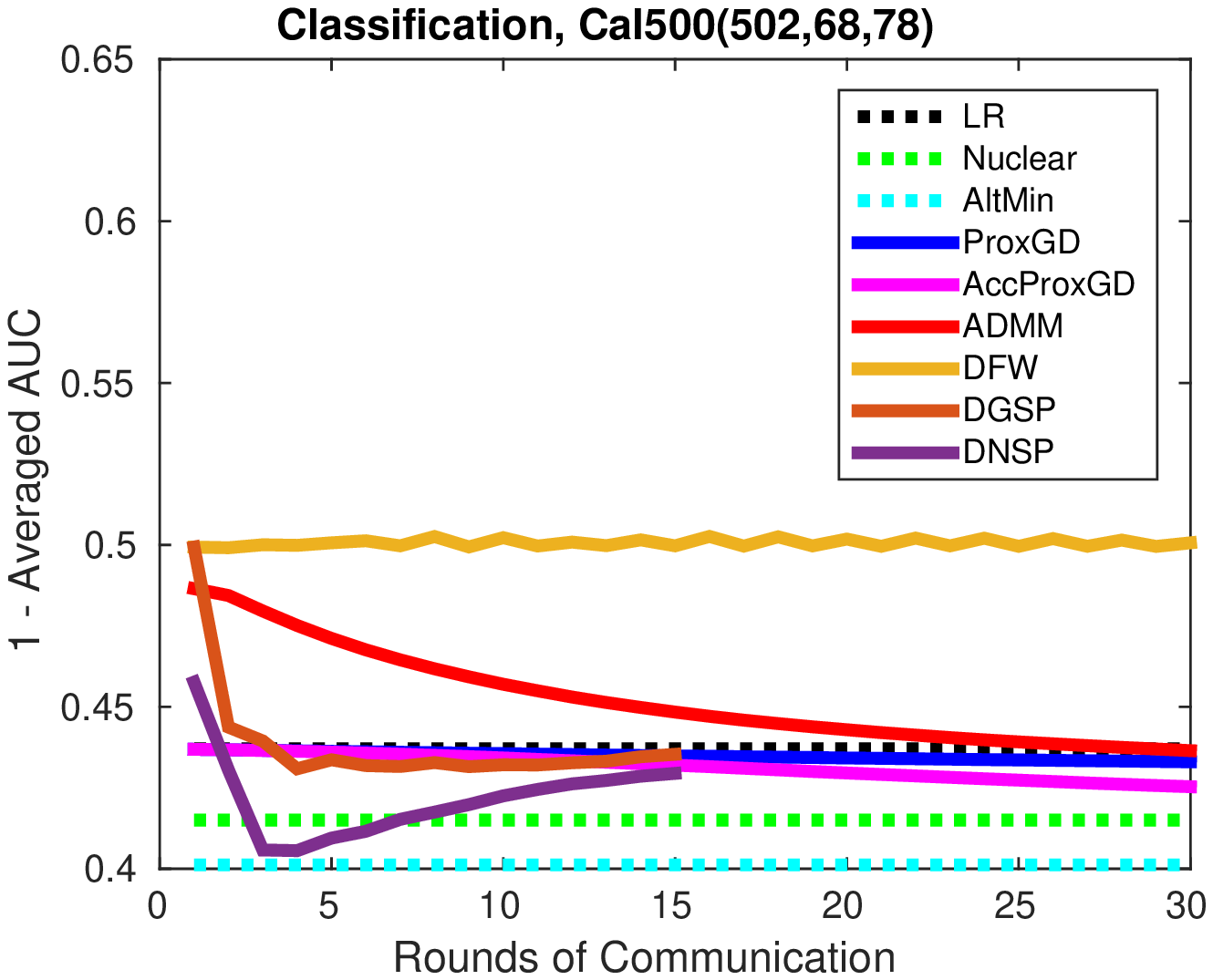}%
\end{center}
\caption{Prediction Error on real data.}
\label{fig:fw_real_data}
\end{figure*}

\end{document}